\documentclass[conference]{IEEEtran}
\usepackage{cite}
\usepackage[pdftex]{graphicx}
\usepackage{algorithm, algorithmic}
\usepackage{amsmath}
\usepackage{array}
\renewcommand{\thispagestyle}[2]{} 

\begin{document}

\title{A Classification Leveraged Object Detector}

\author{\IEEEauthorblockN{Miao Sun}
\IEEEauthorblockA{Electrical and\\Computer Engineering\\
University of Missouri\\
Columbia, Missouri , 65211\\
Email: msqz6@mail.missouri.edu}
\and
\IEEEauthorblockN{Tony X. Han}
\IEEEauthorblockA{Electrical and\\Computer Engineering\\
University of Missouri\\
Columbia, Missouri , 65211\\
Email: hantx@missouri.edu}
\and
\IEEEauthorblockN{Zhihai He}
\IEEEauthorblockA{Electrical and\\Computer Engineering\\
University of Missouri\\
Columbia, Missouri , 65211\\
Email: hezhi@missouri.edu}}

\maketitle

\begin{abstract}

Currently, the state-of-the-art image classification algorithms outperform
the best available object detector by a big margin in terms of average
precision. We, therefore, propose a simple yet principled approach that allows us to leverage
object detection through image classification on supporting regions
specified by a preliminary object detector. Using a simple bag-of-words
model based image classification algorithm, we leveraged the performance of
the deformable model objector from 35.9\% to 39.5\% in average precision over 20 categories on standard PASCAL VOC 2007 detection dataset. 

\end{abstract}

\IEEEpeerreviewmaketitle

\section{Introduction}

To achieve the goal of automatic image understanding, computers should be able
to recognize what objects are in an image and to locate where they are. Image classification is to predict existence of target objects in given images, whereas object detection is to locate each object of a specific class. The location of an object is represented as a bounding box, according to the prestigious and influential PASCAL Visual Object Challenge (VOC)~\cite{Everingham10}. 
Object detection is regarded as more difficult than image classification because object detection requires
predicting not only the presence of each object category but also the
location of each instance. The results of the most recent PASCAL VOC results support
this argument: In terms of average precision (AP), the
winner of the image classification task~\cite{SongCHHY11,Khan12} achieved a mean AP
of $81\%$; the winner of object detection
task~\cite{Felzen08,LongZhuCVPR2010,Vedaldi_ICCV09,vandeSandeICCV2011}  achieved a mean AP below $40\%$~\cite{Everingham12}. This big performance gap forces us to speculate: Can we take advantage of the much better performed image classification algorithm to improve the object detection performance?

Furthermore, available labeled training image data are quite unbalanced for
image classification and object detection. Since most of state-of-the-art
image classification and object detection algorithms are supervised learning
based, the quantity and the quality of the labeled data affect the performance
heavily. This is another reason why we can achieve acceptable performance for
image classification but not for object detection. Besides, We can easily determine the labor
difference between annotating an image for image classification and
annotating an image for object detection: For image classification,
annotators only need to check a list of Yes or No check boxes of relevant object
categories; for object detection, annotators have to label every instance
of each object category with bounding boxes of various scales and aspect
ratios. This labor difference is more salient for large-scale image dataset: In
the standard large scale ImageNet dataset~\cite{imagenet_cvpr09}, there are
$14,197,122$ images of $21,841$ synsets (object categories) labeled for the
image classification task. Among these large numbers of images with categorical
labels, bounding box labels are only available for around $3,000$ popular
synsets, of which the average number of bounding-box labeled images is merely
$150$ image per category~\cite{imagenet_cvpr09}. We can save a huge amount of
human labor if we can train or improve an object detector with image data
labeled for image classification. Therefore, building an image classification leveraged object detector is quite
desirable from the perspective of performance as well as practical application
cost. 

\begin{figure}[t] 
\centering
\includegraphics[width=\linewidth,height=.7\textheight,keepaspectratio]{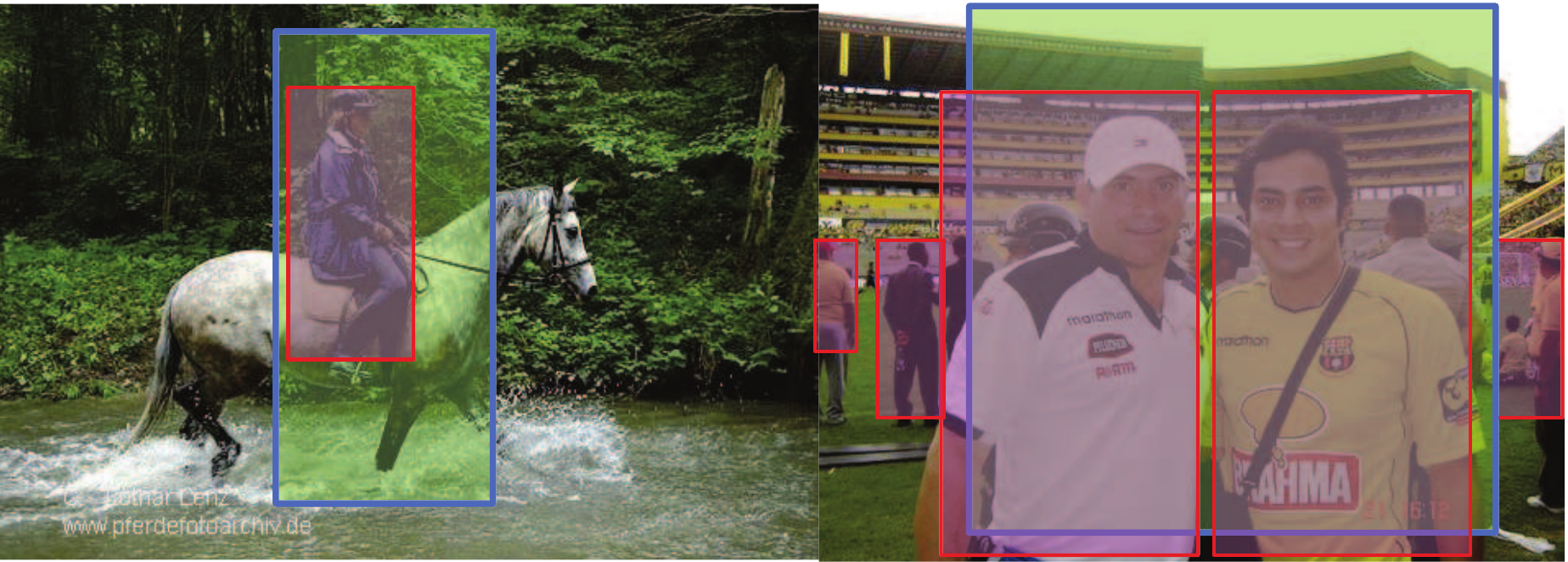}
\caption{\footnotesize{Supporting regions in CLOD. The red
rectangular boxes are detection results from a preliminary object detector. Green regions are created
by subtraction of two boxes. Both the magenta regions and green regions are called supporting
regions, which will be the input for classification algorithm.}}\label{fig:idea}
\end{figure}

\begin{figure*}[t] 
\centering
\includegraphics[width=\linewidth,height=.7\textheight,keepaspectratio]{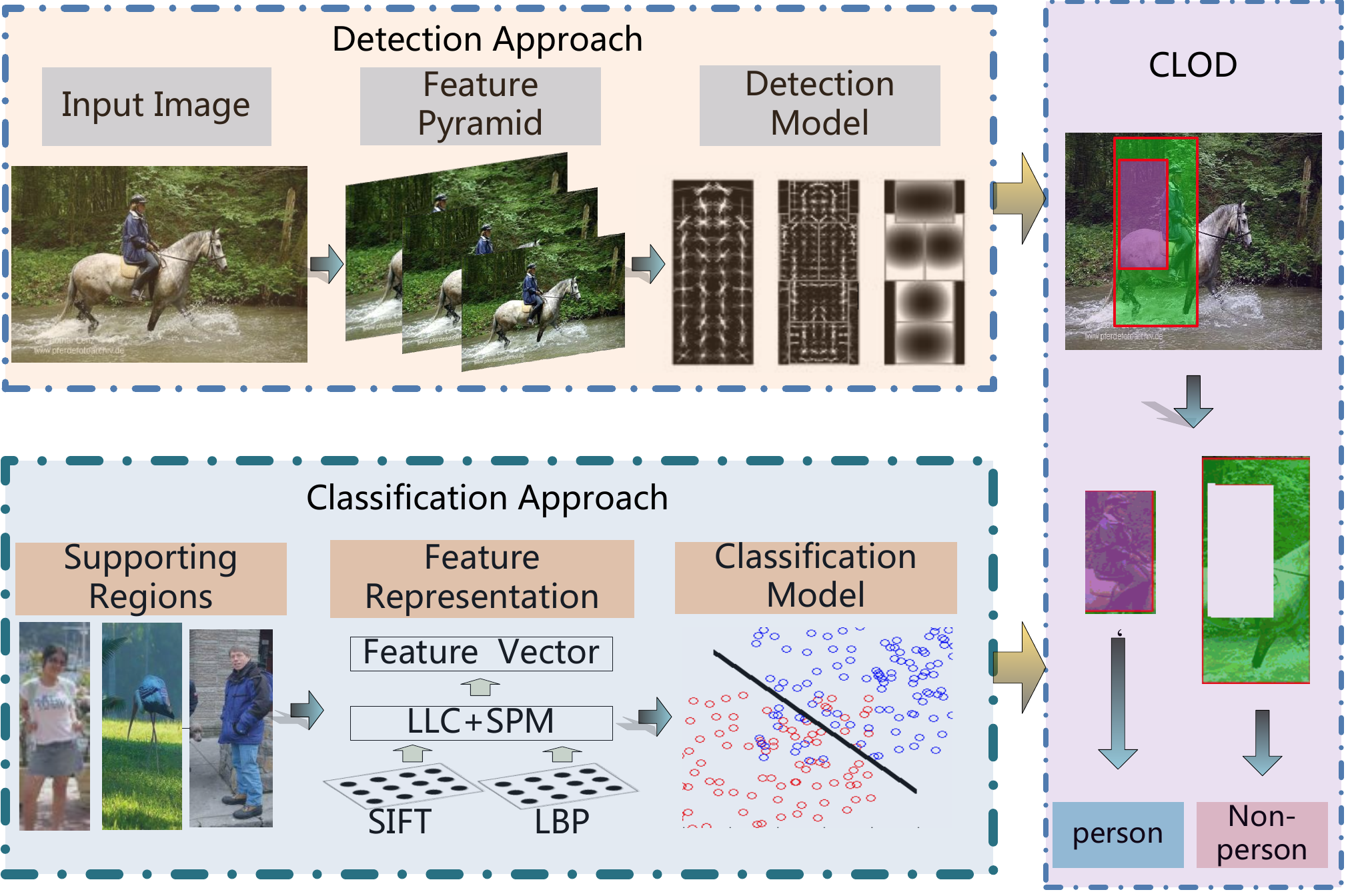}
\caption{Workflow for classification leveraged Object Detector. Given the ground truth bounding boxes, we train one DPM detector as it is shown in the first row and we train one classification classifier as it is show in the second row. The supporting regions are obtained via subtraction of the detection bounding boxes, then the support regions are feed into the classification classifier to further predict the presence of target object}\label{fig:framework}
\end{figure*}

However, there are several factors we need to consider in order to apply the
available image classification algorithms to object detection. {\it First}, simply
applying the state-of-the-art image classification
algorithms~\cite{Grauman05thepyramid,Lazebnik06,Boiman08,BoschZM08,Zhou09,WangJ10,SongCHHY11,YangN11,Zhang2006}
to each scanning window is not feasible due to the speed issue. Most of the
aforementioned image classification algorithms
~\cite{Grauman05thepyramid,Lazebnik06,Zhou09,WangJ10,SongCHHY11,YangN11} use
one or several key classic components including the Bag of Words(BOW) model of a large size
codebook, Spatial Pyramid Matching(SPM), and feature pooling, which makes
feature extraction very slow compared with the modern sliding-window-based
detectors~\cite{Dalal05,Felzen_pami10,LongZhuCVPR2010,DesaiRF11}. Usually a
sliding-window-based object detector will scan hundreds of thousands of sliding
windows in order to detect every instances in the image. If we directly apply
image classification algorithms to each scanning window, object detection in an
image is equivalent to classifying hundreds of thousands of images. {\it Second},
if we apply image classification to selected candidate regions as in ~\cite{vandeSandeICCV2011}, i.e.\ , the selective search on oversegmented
superpixels, the image classification algorithm should be robust to region
cropping and clipping and should remain discriminative.

We propose a simple yet principled approach to leveraging object
detection through image classification on supporting regions specified by a
preliminary object detector. Using a simple bag-of-words model based image
classification algorithm, we leveraged the performance of the deformable model objector from 35.9\% to 39.5\% in average precision, which led to the state-of-the-art performance on the standard PASCAL VOC 2007 detection dataset. 

%

\section{Classification Leveraged Object Detector}
\label{sec:clod}

A Classification Leveraged Object Detector(CLOD) mainly contains three components: detection algorithm, classification algorithm and supporting regions for classification. We will first discuss the details about the supporting regions in section~\ref{sec:region}, then CLOD will be proposed to incorporate supporting regions, detection algorithm and classification algorithm into a cascade style framework in section ~\ref{sec:workflow}.

\subsection{Supporting Regions}
\label{sec:region}
Object detection can be formulated as a classification algorithm if given the possible
bounding boxes, which indicates the possible object locations. However, directly applying the classification algorithm to bounding boxes will result in bad performance , because the multiple detection boxes of the same object in an image will considered false detections. Therefore, we are going to introduce supporting regions concept for classification algorithm. Some supporting regions examples are illustrated in Figure~\ref{fig:idea}.  Some possible locations are predicted by a preliminary object detector, then the supporting regions are generated based on the predicted bounding boxes according the formulation below:

Let $\overline{D}_i$ be the
detection candidate boxes, $i$ is from 1 to N for a single image. We sort the
boxes so that the detection score of $\overline{D}_i$ is larger than
$\overline{D}_j$, if $i < j$. Let $B$ be the background region, which can be expressed as

\begin{equation}
	B=\bigcap_{i=1}^{N} \overline{D}_i
\end{equation}
If there are no missing detections, the classification score of $B$ would satisfy
\begin{equation}
	f_c(B) < 0
\end{equation}

Then $i$ is from 1 to N , that is, the boxes we want to classify are from high detection score to low detection score.

\begin{align}
	S_k &= B\cup \big(\bigcup_{i > k}(D_k\cap \overline{D}_i)\big) \\
	    &= B\cup \big(D_k-\bigcup_{i > k}(D_k\cap D_i)\big) 
\end{align}

This equation means the classification region for detection box $k$ will only be
affected by the the boxes whose detections scores are higher .

As previously mentioned, there may be mis-detections in the image, which will
affect our results a lot. So we define a background region as follows:

\begin{equation}
	{B}_i={D}_i^{c}\cap (\bigcap_{i=1}^{N} \overline{D}_i) - {D}_i
\end{equation}

In this equation, ${D}_i^{c}$ is the box ${D}_i$ with an extra margin. 


\subsection{Workflow for Classification Leveraged Object Detector }
\label{sec:workflow}

\begin{table*}[t]  {
  \caption{Classification Performance on PASCAL VOC 2007 Classification Dataset.}
  \label{table:Cls07}}
 \centering
 \setlength{\tabcolsep}{3pt}
 \begin{tabular}{|c||c|c|c|c|c|c|c|c|c|c|c|c|c|c|c|c|c|c|c|c|c|}
\hline
\;\; &plane&bike&bird& boat& bottle& bus& car& cat& chair& cow
\\\hline
INRIA Genetic~\cite{inriaG} &77.5&63.6&56.1&71.9&33.1&60.6&78.0&58.8&53.5&42.6
\\\hline
SuperVec~\cite{superVec}      &79.4&72.5&55.6&73.8&34.0&72.4&83.4&63.6&56.6&52.8
\\\hline
INRIA 2009~\cite{locCls}    &77.2&69.3&56.2&66.6&45.5&68.1&83.4&53.6&58.3&51.1
\\\hline
TagModal~\cite{TagModal}      &87.9&65.5&76.3&75.6&31.5&71.3&77.5&79.2&46.2&62.7
\\\hline
CODC~\cite{CODC}          &82.5&79.6&64.8&73.4&54.2&75.0&87.5&65.6&62.9&56.4
\\\hline
SIFT+LLC &73.1&61.2&49.1&65.5&26.0&55.0&75.7&56.9&51.7&36.1
\\\hline
LBP+LLC  &74.8&54.3&40.7&65.1&20.9&53.0&69.9&54.8&50.7&31.8
\\\hline
SIFT+LBP+LLC&77.2&64.3&52.7&70.4&27.2&60.3&77.3&61.0&54.6&40.2
\\\hline
\;\; & table& dog& horse& motor& person& plant& sheep& sofa& train& tv& mAP
\\\hline
INRIA Genetic~\cite{inriaG} &54.9&45.8&77.5&64.0&85.9&36.3&44.7&50.6&79.2&53.2&59.4
\\\hline                                                        
SuperVec~\cite{superVec} &63.2&49.5&80.9&71.9&85.1&36.4&46.5&59.8&83.3&58.9&64.0
\\\hline                                                        
INRIA 2009~\cite{locCls} &62.2&45.2&78.4&69.7&86.1&52.4&54.4&54.3&75.8&62.1&63.5
\\\hline                                                          
TagModal~\cite{TagModal} &41.4&74.6&84.6&76.2&84.6&48.0&67.7&44.3&86.1&52.7&66.7
\\\hline                                                        
CODC~\cite{CODC} &66.0&53.5&85.0&76.8&91.1&53.9&61.0&67.5&83.6&70.6&70.5
\\\hline                                                        
SIFT+LLC&46.8&39.5&76.1&61.9&81.6&25.5&42.3&52.2&73.9&50.25&55
\\\hline                                                        
SIFT+LLC&40.8&42.6&72.9&46.8&80.3&22.2&34.8&43.7&72.7&39.06&50.6
\\\hline                                                        
SIFT+LBP+LLC&53.8&46.9&77.2&62.4&84.0&26.8&44.1&54.2&77.2&51.4&58.2
\\\hline

\end{tabular}
\end{table*}

\begin{table*}[t]  {
\caption{ Comparison of different types of classification classifiers.}
 \label{table:clod_lbp}}
 \centering
 \setlength{\tabcolsep}{3pt}
 \begin{tabular}{|c||c|c|c|c|c|c|c|c|c|c|c|c|c|c|c|c|c|c|c|c|c|}
\hline
\;\; &plane&bike&bird& boat& bottle& bus& car& cat& chair& cow
\\\hline
Det     &35.7&59.8&11.8&19.6&31.0&51.8&58.7&29.3&23.4&28.7
\\\hline
CLS-I   &36.4&59.8&11.8&19.6&31.0&51.8&58.8&29.3&23.6&28.7
\\\hline
CLS-Rg   &37.0&60.1&12.2&20.6&31.9&53.4&59.6&32.3&24.0&31.4
\\\hline
CLS-Rf  &36.5&59.9&12.0&20.0&31.1&52.3&58.7&30.4&23.5&29.5
\\\hline
\;\; & table& dog& horse& motor&person& plant& sheep& sofa& train& tv& mAP
\\\hline
Det        &26.0&15.5&60.1&50.5&44.1&13.3&27.7&37.6&48.8&45.3&35.9
\\\hline                                                      
CLS-I   &26.0&15.5&60.1&50.5&44.1&13.5&27.7&37.6&48.8&45.3&36.0
\\\hline                                                      
CLS-Rg    &29.8&17.2&61.7&53.0&44.4&15.1&27.8&40.6&49.8&45.3&37.3
\\\hline                                                      
CLS-Rf  &26.6&16.5&60.6&51.0&44.2&14.4&27.7&37.8&48.9&45.7&36.3
\\\hline

\end{tabular}
\end{table*}

Our detection algorithm is a enhanced deformable part models, where we developed a new feature called PCA-reduced-HOG-LBP feature. First, each input region is split into $16\times16$ blocks. Both HOG and LBP features are extracted from all the blocks. Then we concatenate the HOG and LBP feature for each block and apply pca on them. The final concatenated PCA-reduced-HOG-LBP is incorporated into deformable part models framework. 

Our classification algorithm has following steps: feature extraction, coding , pooling and classification. In this paper, we use dense SIFT and LBP features, and adopt Locality-constrained Linear coding(LLC) to enhance the feature representation. Then Spatial Pyramid Matching(SPM) is applied to make the feature more robust. The final feature representation is sent to linear SVM to predict the presence of the target object in an image.

Given detection algorithm, classification algorithm and supporting regions, the workflow for our classification leveraged object detector is described in Figure~\ref{fig:framework}. First, we apply detection approach. The we could get some predicted bounding boxes. Then, the supporting regions are generated according to the rules in Section~\ref{sec:region}. We train one classification classifier using those supporting regions from training data and apply the classification model to these supporting regions from testing data. 

we implemented a simple but powerful procedure to boost the performance based on the detection results and classification scores:  
Let $(D_1 , . . . , D_k )$ be a set of detections obtained using $k$ different object categories in an image $I$. 
Each detection $D_i=(B, s)$ is defined by a bounding box $B = (x_1 , y_1 , x_2 , y_2 )$ and a confidence score $s$. Detection score information is defined as $ f_1(I) = (\alpha(s_1), . . . , \alpha(s_k ))$ where $s_i$ is the score of the highest scoring detection in $D_i$ , and $\alpha(x) = 1/(1 + exp(−2x))$ is a logistic function for renormalizing the scores. Classification score information $ f_2(I) $ is defined in similar procedure. So the final combined feature for each box is a $2k+6$ dimensional feature:  $[\alpha(d_i), \alpha(c_i), x_1, y_1, x_2, y_2, f_1(I), f_2(I)]$. Finally, SVM with RBF kernel is applied to rescore the preliminary detection confidence scores.

\section{Experiments and Discussion}

To demonstrate the advantage of our approach, we test the CLOD on the very challenging PASCAL Visual Object Challenge 2007 (VOC2007) datasets~\cite{Everingham10} . First, we give a detailed description of VOC2007 dataset and the cropped dataset for our CLOD framework.
Then, we evaluate our classification algorithm on PASCAL VOC2007 classification dataset. After that we compare the CLOD performance with the state the art detection performance on PASCAL VOC2007 detection dataset.

\subsection{Datasets and Metrics}

\subsubsection{Datasets}
\label{sec:region_set}

PASCAL VOC2007  datasets ~\cite{Everingham10} has 20 categories, containing 9,963 images and 24,640 objects. This dataset is divided into ``train'', ``val''and ``test'' subsets, which contains 2501, 2510 and  4592 images respectively. Parameters of the algorithm are tuned via training on ``train'' set and evaluating on ``val'' set. The final model is trained on ``train'' + ``val'' sets and is applied on the ``test'' set to obtain the final results.  The dataset is extremely challenging since the objects vary significantly in size, view angle, illumination, appearance and pose. 

Notice the the classification is applied on the supporting regions instead of the the whole image as it is shown in Section~\ref{sec:region}, so a region level (achieved by cropping the bounding boxes from the dataset) seems necessary for a satisfactory classification classifier in CLOD.  We prepare a region-level dataset by cropping the detection ground truth boxes according the detection annotations. This cropped dataset contains ``train'' and ``val''.  The positive examples are the ground truth bounding boxes, while the negative examples are the ground truth bounding boxes from other categories. ``test'' subset is not used to for region-level classification classifier. We refer this region-level dataset as pure ground-truth-box set. 

There is another way to get classification classifiers: We can use detection false alarm boxes as the negative and the ground truth boxes as the positive. Notice that the false alarms boxes here are applied to the supporting region technique, so that the false alarms do not have any part of the ground truth. In this way, each category has a different kmeans codebook, while features from other categories are not taken into consideration. For this task, the positive examples are the sum of ground truth in ``train'' and ``val'' ,  and negative examples are a random selection from the false alarms from the detection boxes in the ``trainval'' dataset. We refer this region-level dataset as ground-truth-false-alarms set.

\begin{table*}[tp] { 
 \caption{ {Comparison with the state-of-the-art performance of object detection on PASCAL VOC 2007.}}
 \label{table:state_of_the_art}}
 \centering
 \setlength{\tabcolsep}{3pt}
 \begin{tabular}{|c||c|c|c|c|c|c|c|c|c|c|c|c|c|c|c|c|c|c|c|c|c|}
\hline
\;\; &plane&bike&bird& boat& bottle& bus& car& cat& chair& cow
\\\hline
Leo~\cite{LongZhuCVPR2010}                 &29.4&55.8&9.4&14.3&28.6&44.0&51.3&21.3&20.0&19.3 
\\\hline
CMO~\cite{LiPC11}                          &31.5&61.8&12.4&18.1&27.7&51.5&59.8&24.8&23.7&27.2
\\\hline
INRIA2009                                  &35.1&45.6&10.9&12.0&23.2&42.1&50.9&19.0&18.0&31.5
\\\hline
UoC2010                                    &31.2&61.5&11.9&17.4&27.0&49.1&59.6&23.1&23.0&26.3
\\\hline
Det-Cls~\cite{SongCHHY11}                  &38.6&58.7&{\underline{\bf 18.0}}&18.7&31.8&53.6&56.0&30.6&23.5&31.1
\\\hline
Oxford~\cite{Vedaldi_ICCV09}               &37.6&47.8&15.3&15.3&21.9&50.7&50.6&30.0&17.3&33.0
\\\hline
NLPR~\cite{ChineseAcademy2010or2011}       &36.7&59.8&11.8&17.5&26.3&49.8&58.2&24.0&22.9&27.0
\\\hline
Ver.5~\cite{voc-release5}                  &36.6&62.2&12.1&17.6&28.7&54.6&60.4&25.5&21.1&25.6
\\\hline
MOCO ~\cite{Guang13}                      &{\underline{\bf 41.0}}&{\underline{\bf 64.3}}&15.1&19.5&33.0&{\underline{\bf 57.9}}&{\underline{\bf 63.2}}&27.8&23.2&28.2
\\\hline
CLOD                        &38.9 &62.4 &16.5 &{\underline{\bf22.7}} &{\underline{\bf 32.2}} &54.8 &60.9 &{\underline{\bf 34.0}} &{\underline{\bf 25.4}}&{\underline{\bf 33.4}} 
\\\hline
\;\; & table& dog& horse& motor&person& plant& sheep& sofa& train& tv& mAP
\\\hline
Leo~\cite{LongZhuCVPR2010}                 &25.2&12.5&50.4&38.4&36.6&15.1&19.7&25.1&36.8&39.3&29.6
\\\hline                                                                                          
CMO~\cite{LiPC11}                          &30.7&13.7&60.5&51.1&43.6&14.2&19.6&38.5&49.1&44.3&35.2
\\\hline                                                                                         
INRIA2009                                  &17.2&17.6&49.6&43.1&21.0&{\underline{\bf 18.9}}&27.3&24.7&29.9&39.7&28.9
\\\hline                                                                                          
UoC2010                                    &24.9&12.9&60.1&51.0&43.2&13.4&18.8&36.2&49.1&43.0&34.1
\\\hline                                                                                          
Det-Cls~\cite{SongCHHY11}                  &{\underline{\bf 36.6}}&20.9&62.6&47.9&41.2&18.8&23.5&41.8&{\underline{\bf 53.6}}&45.3&37.7
\\\hline                                                                                          
Oxford~\cite{Vedaldi_ICCV09}               &22.5&{\underline{\bf 21.5}}&51.2&45.5&23.3&12.4&23.9&28.5&45.3&48.5&32.1
\\\hline                                                                                          
NLPR~\cite{ChineseAcademy2010or2011}       &24.3&15.2&58.2&49.2&44.6&13.5&21.4&34.9&47.5&42.3&34.3
\\\hline                                                                                          
Ver.5~\cite{voc-release5}                  &26.6&14.6&60.9&50.7&44.7&14.3&21.5&38.2&49.3&43.6&35.4
\\\hline                                                                                          
MOCO~\cite{Guang13}                       &29.1&16.9&63.7& 53.8&{\underline{\bf 47.1}}& 18.3& 28.1& 42.2& 53.1&{\underline{\bf 49.3}}&38.7
\\\hline                                                                                          
CLOD                           &34.2 &20.0 &{\underline{\bf 63.8}} &{\underline{\bf 55.1}} &45.7 &18.6 &{\underline{\bf 30.4}} &{\underline{\bf 42.6}} &51.4 &47.8 &{\underline{\bf39.5}}

\end{tabular}

\end{table*}

\subsubsection{Metrics}
{\em Average Precision (AP)} For the VOC2007 Challenge, the interpolated average precision ~\cite{Salton86} was used to evaluate both classification and detection. 

For a given task and class, the precision/recall curve is
computed from a method’s ranked output. Recall is defined
as the proportion of all positive examples ranked above a
given rank. Precision is the proportion of all examples above
that rank which are from the positive class. The AP summaries the shape of the precision/recall curve, and is defined as the mean precision at a set of eleven equally spaced
recall levels $[0, 0.1, . . . , 1]$:
\begin{equation}
  AP=\frac{1}{11} \sum_{r \in {0, 0.1, ..., 1}} P_{interp}(r)
\end{equation}

The precision at each recall level $r$ is interpolated by taking the maximum precision measured for a method for which the corresponding recall exceeds r:
\begin{equation}
  P_{interp}(r)=\max_{\hat{r}:\hat{r} \geq r} p(\hat{r})
\end{equation}
Where $p(\hat{r})$ is the measured precision at recall $\hat{r}$

{\em Bounding Box Evaluation} As noted, for the detection task,
participants submitted a list of bounding boxes with associated score (rank). Detections were assigned to ground truth objects and judged to be true/false positives by measuring bounding box overlap. To be considered a correct detection, the overlap ratio $a_o$ between the predicted bounding
box $B_p$ and ground truth bounding box $B_gt$ must exceed 0.5
(50\%) by the formula,
\begin{equation}
  a_o = \frac{B_p \cap B_gt}{B_p \cup B_gt}
\end{equation}
where $B_p \cap B_gt$ denotes the intersection of the predicted and ground truth bounding boxes and $B_p \cup B_gt$ their union.

\subsection{Classification Classifier}
In this section, we first tune the parameters of our classification algorithm using the image-level dataset and compare the performance with other state-of-the-art classification algorithms. 
Then we fix those parameters and apply the classification algorithm in our CLOD framework to compare the image-level classifier and region-level classifier.

\subsubsection{Image-level classification}
For our classification method, we choose dense SIFT and LBP as features and BoF+SPM+LLC system. For both dense SIFT and LBP, we adopt a multi-scale technique, in which the patch size for dense SIFT is $8\times8, 16\times16, 25\times25, 36\times36$  and the patch size for LBP is $12\times12, 16\times16, 20\times20, 24\times24$. The stride for dense SIFT is 4 and LBP is 50\% overlap stride. After extracting the dense SIFT and LBP features, a codebook is trained separately by kmeans.
The codebook size for each feature is 10,240 and the spatial pyramid matching is using $1\times1, 1\times2$, and $2\times3$. Therefore, each image would have a 184,320-dimension feature. We can see the performance of our classification classifier on PASCAL VOC 2007 and compare it with the other state-of-the-art classification algorithms in Table~\ref{table:Cls07}. 

From Table~\ref{table:Cls07}, we could see that our classifier is not the best one, but later we will prove that even with this below-average classification classifier, our CLOD approach would still be able to boost the detection a lot.

\subsubsection{Region-level Classification}

From Section~\ref{sec:region_set}, there are two kinds of region-level dataset.With the exact same experiment setup, we train our region-level classifier on the ``train'' + ``val'' subsets of the cropped ground truth dataset and support-region dataset. We evaluate it on the ``test'' subset, the performance is listed in Table~\ref{table:clod_lbp}. In Table~\ref{table:clod_lbp}, only the LBP feature is used due to the speed issue. CLS is adding classification score to the original detection score instead of the complex rescore scheme in Section~\ref{sec:workflow}. Det means the performance of preliminary detection results. CLS-I means CLS using classification classifiers trained on image-level set. CLS-Rg means CLS using classification classifiers trained on region-level pure ground-truth-box set. 

We can see that the image-level classifier is the worst and the region-level classifier from the pure ground-truth-box set is the best. Since our CLOD  actually applied the classification on the supporting regions instead of the whole images, it is reasonable that the image-level classifier does not work well. But it is quite interesting that the classifier from the pure ground-truth-box dataset is better than the classifier from ground-truth-false-alarms set. The reason behind this results is that negative examples from ground truth of other categories carry more discriminative information than the negative examples from false alarms of detection detector.
Therefore, by taking all of the above into consideration, we choose the classification classifier from the pure-ground-truth dataset as our classification classifier in our CLOD framework.

\subsection{Leverage Detection with Classification}

Now, we have already discussed the datasets and details for the CLOD framework. We have showed that each box has a detection score and classification score. Notice that the classification score is not achieved by the whole bounding box region but the supporting region. 

From Section~\ref{sec:region}, the supporting regions are defined as the subtractions of bounding boxes from detection classifiers. In fact given different detection threshold, there will be different number of detection bounding boxes. To reduce the candidate boxes for each class, we set threshold to -0.95 for all the categories, which will lead most categories to contain less than 2,000  candidate boxes.  

\begin{figure*}[t]
  \centering
  \begin{tabular}{cc} 
  \includegraphics[width=35mm]{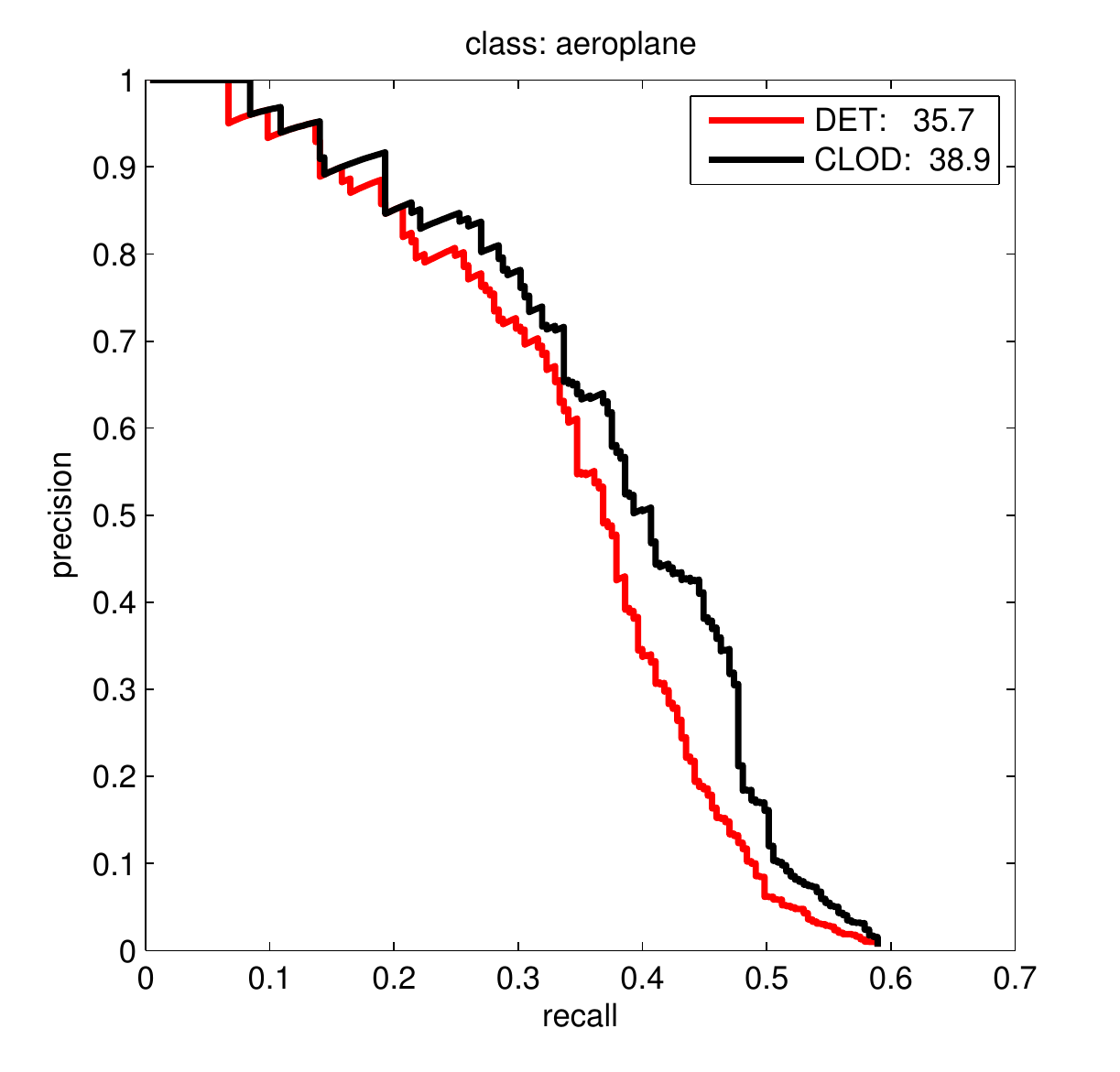} 
  \includegraphics[width=35mm]{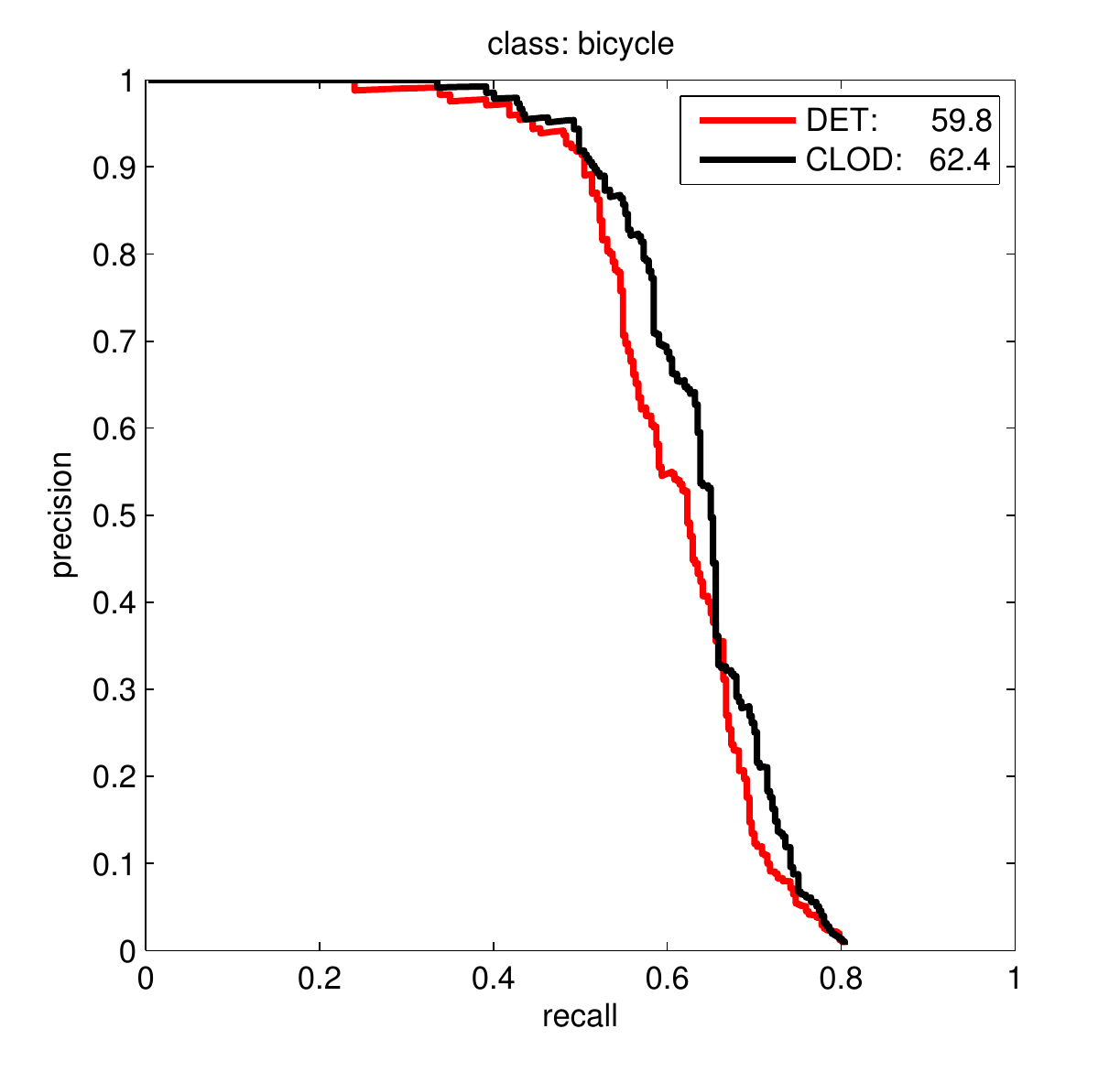}  
  \includegraphics[width=35mm]{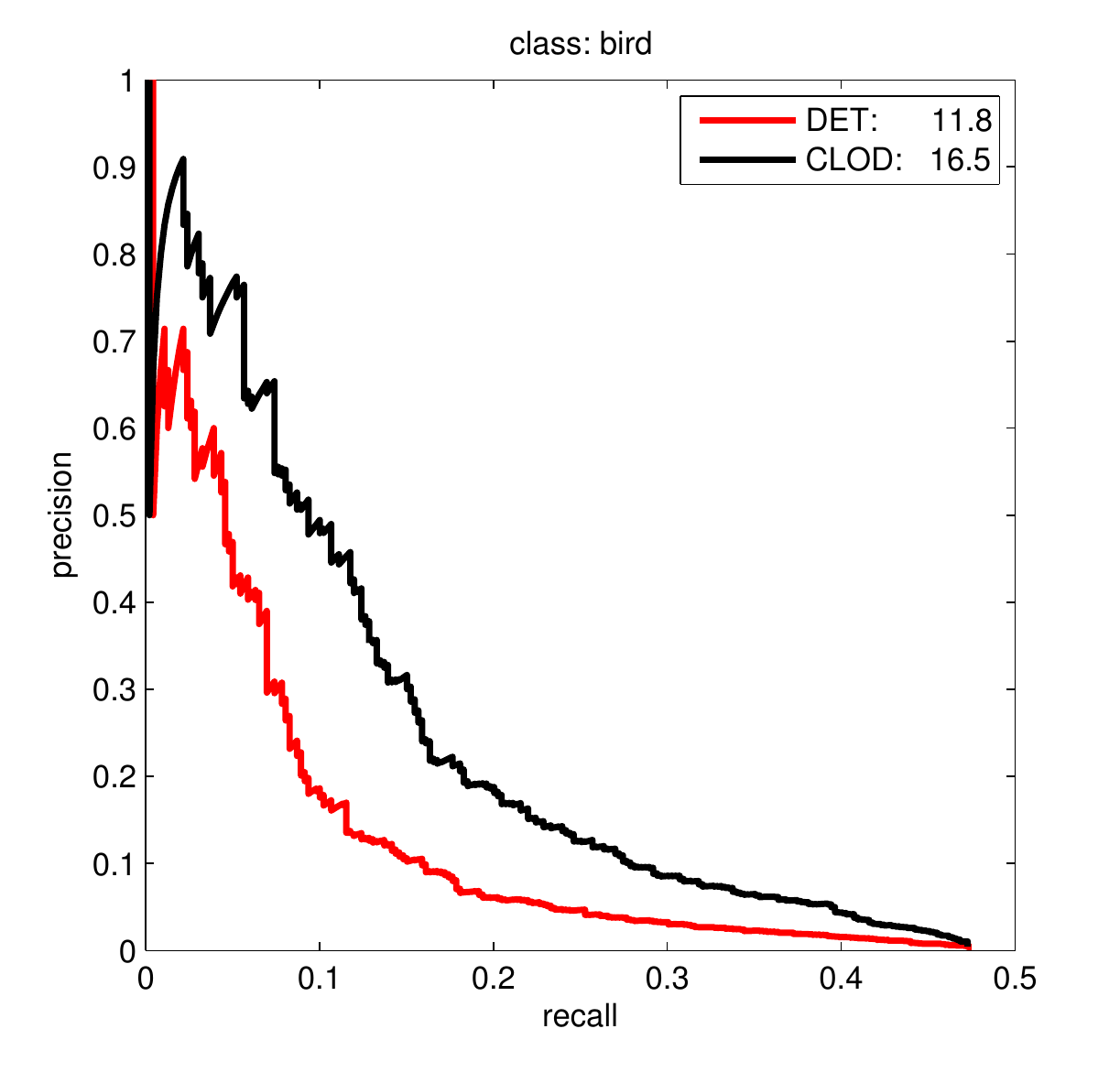}  
  \includegraphics[width=35mm]{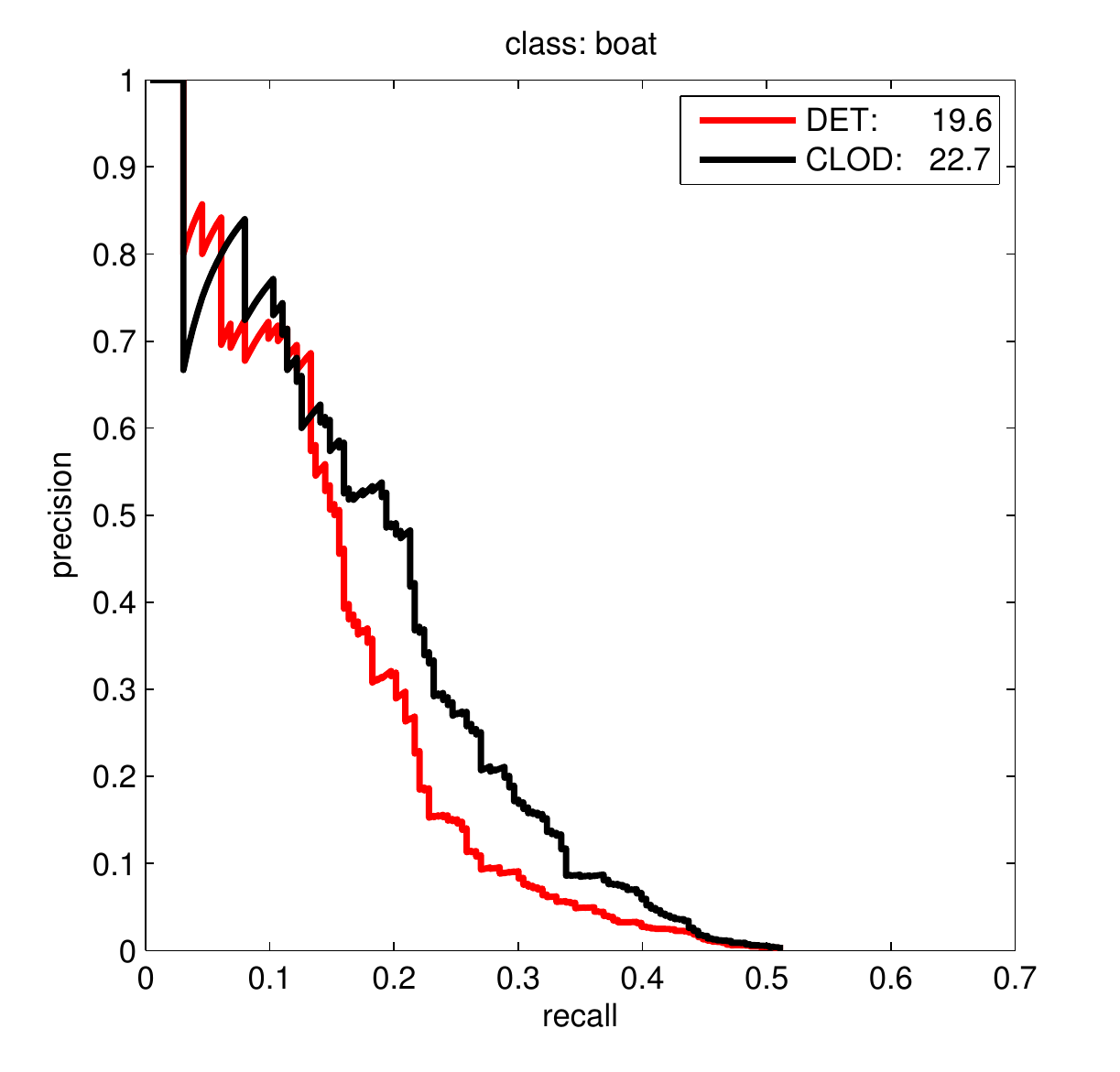}
  \includegraphics[width=35mm]{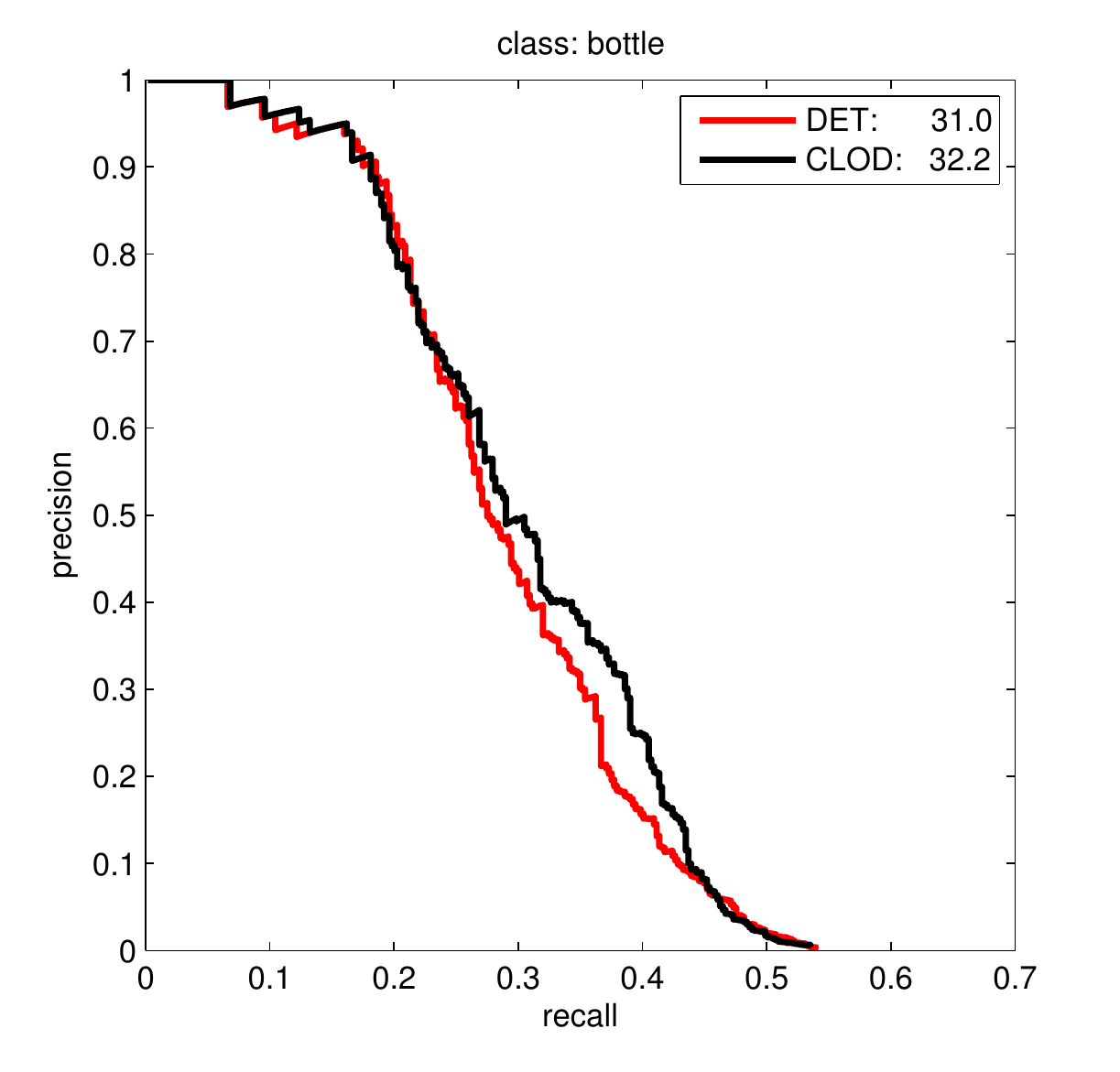} \\   
  \includegraphics[width=35mm]{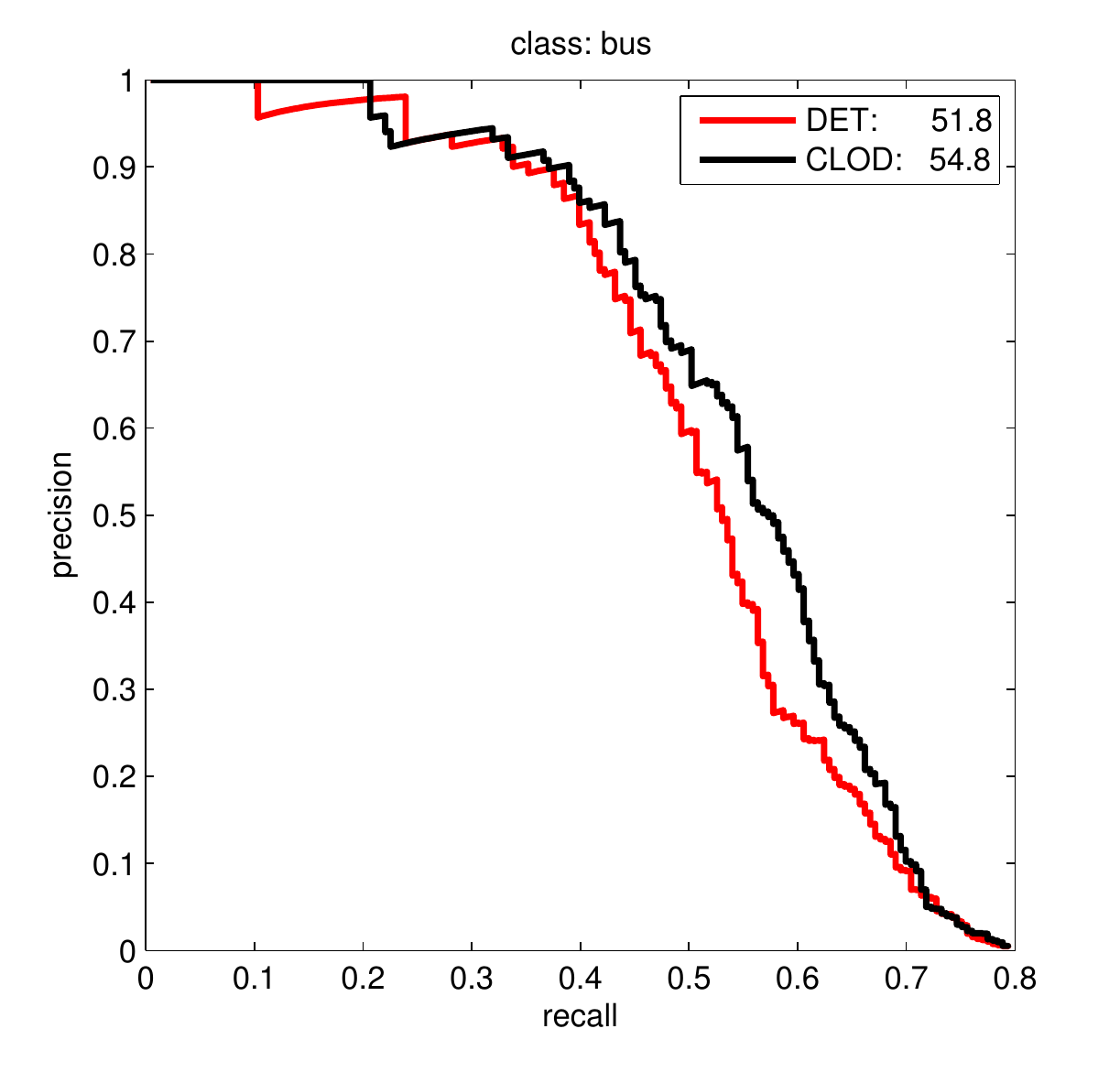}
  \includegraphics[width=35mm]{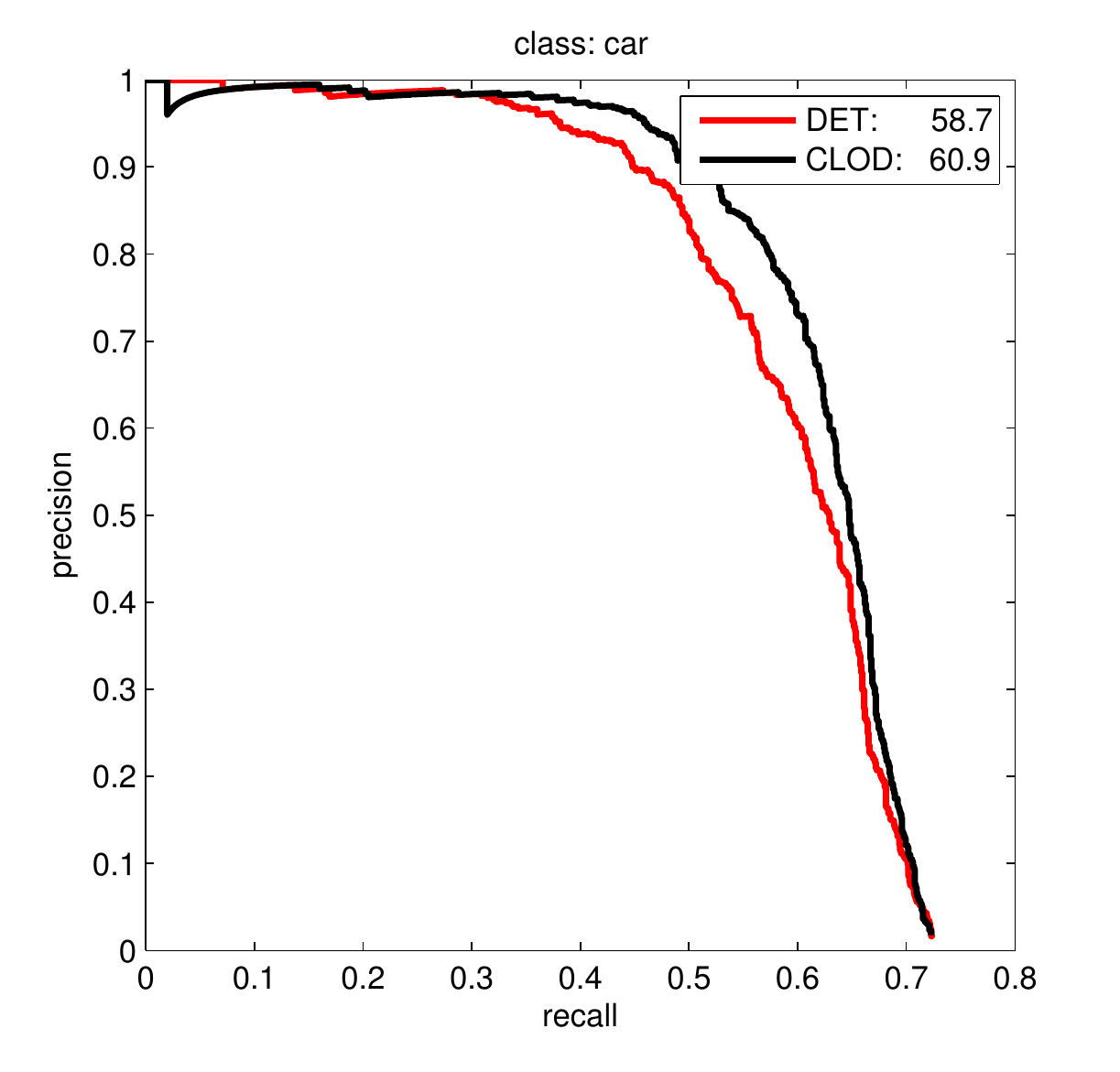}
  \includegraphics[width=35mm]{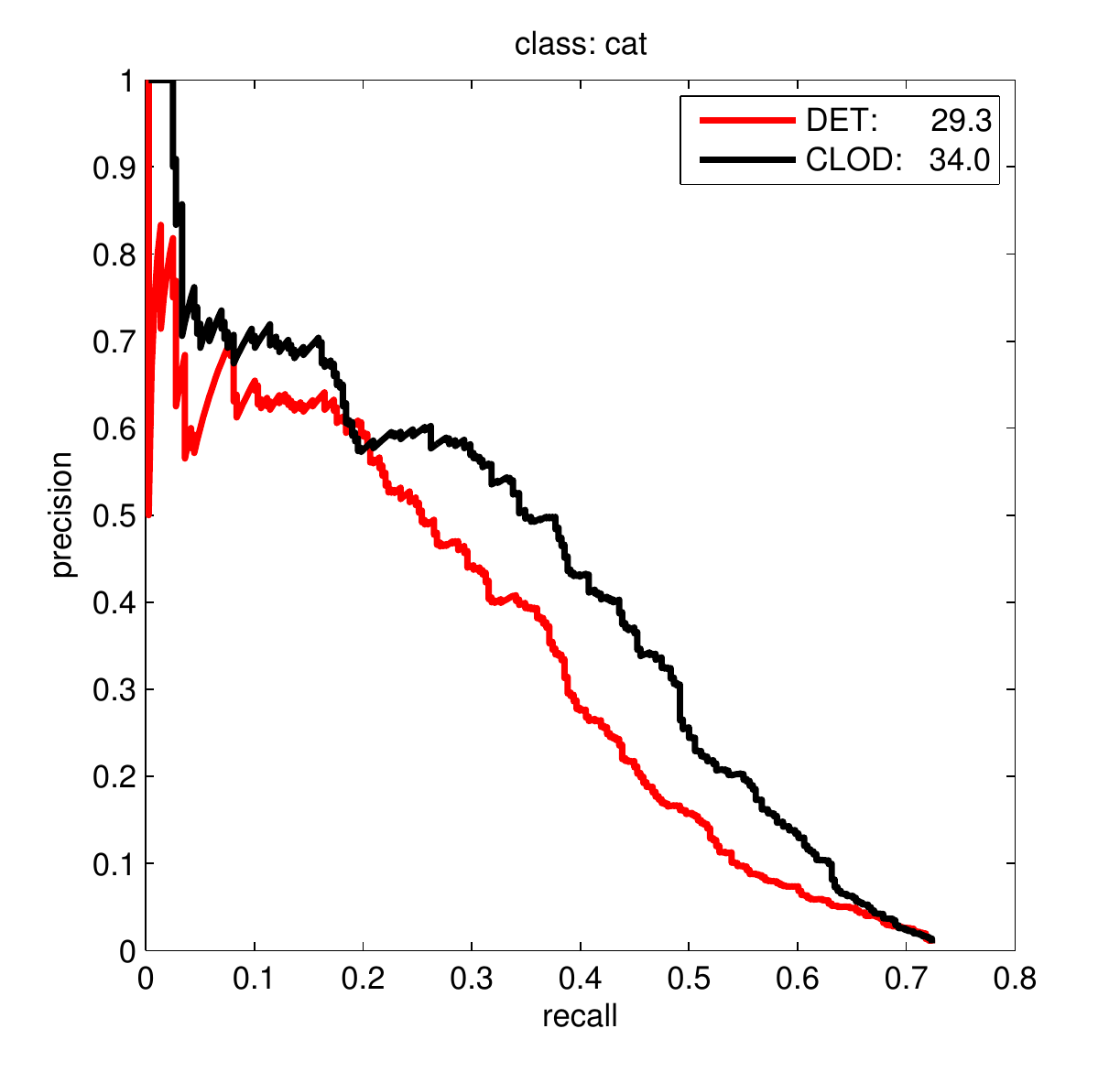}
  \includegraphics[width=35mm]{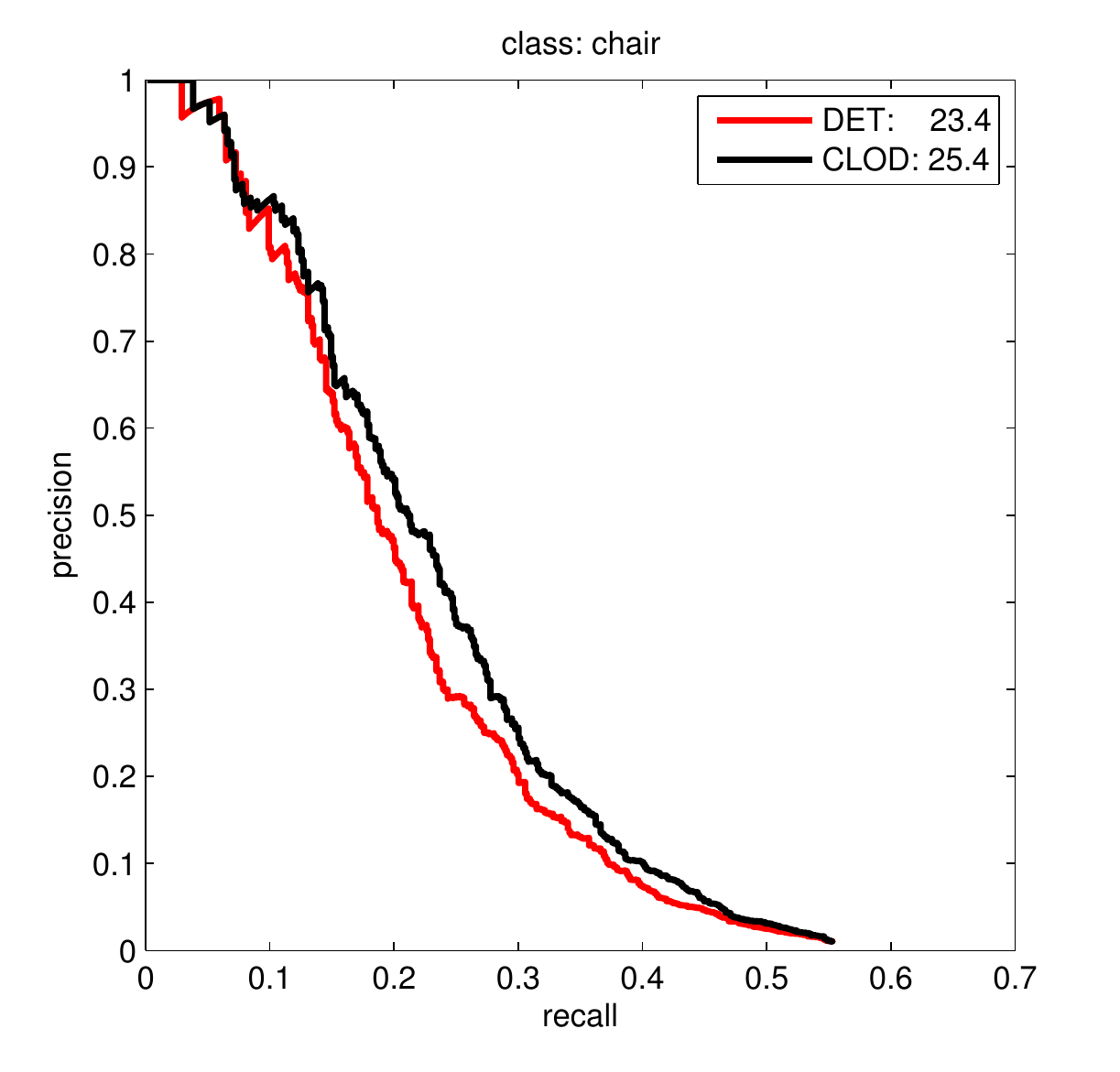}
  \includegraphics[width=35mm]{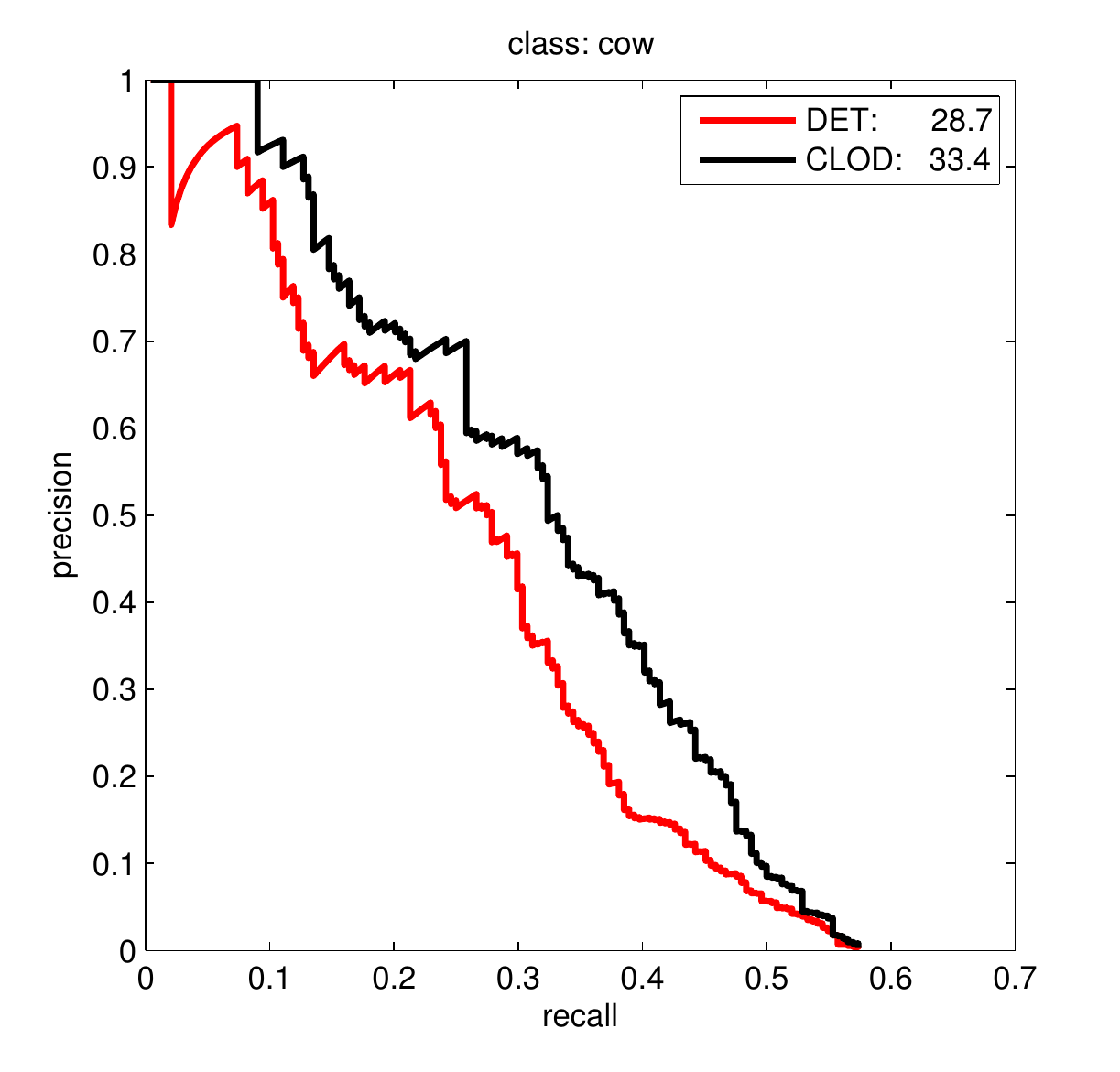} \\
  \includegraphics[width=35mm]{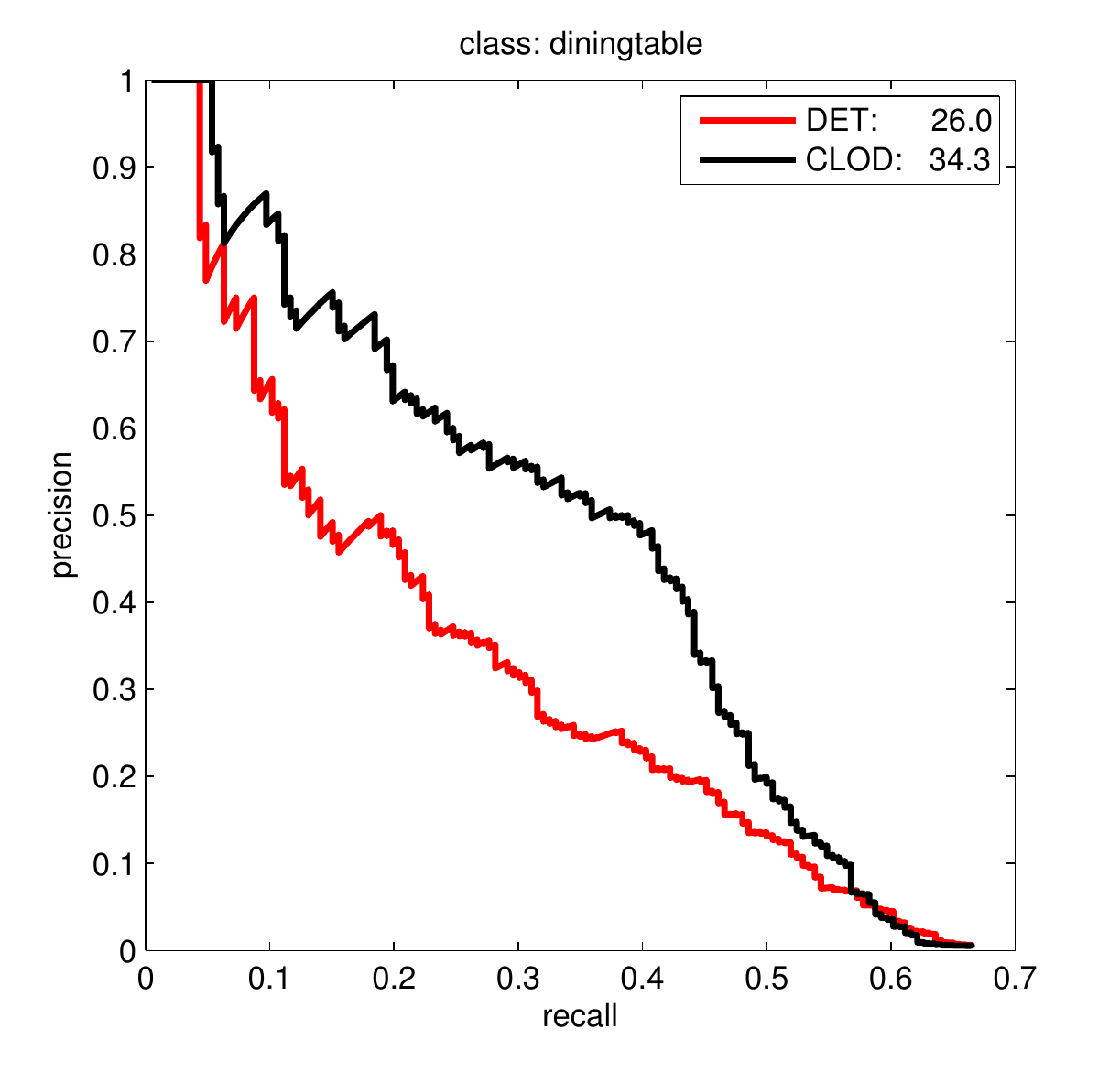}  
  \includegraphics[width=35mm]{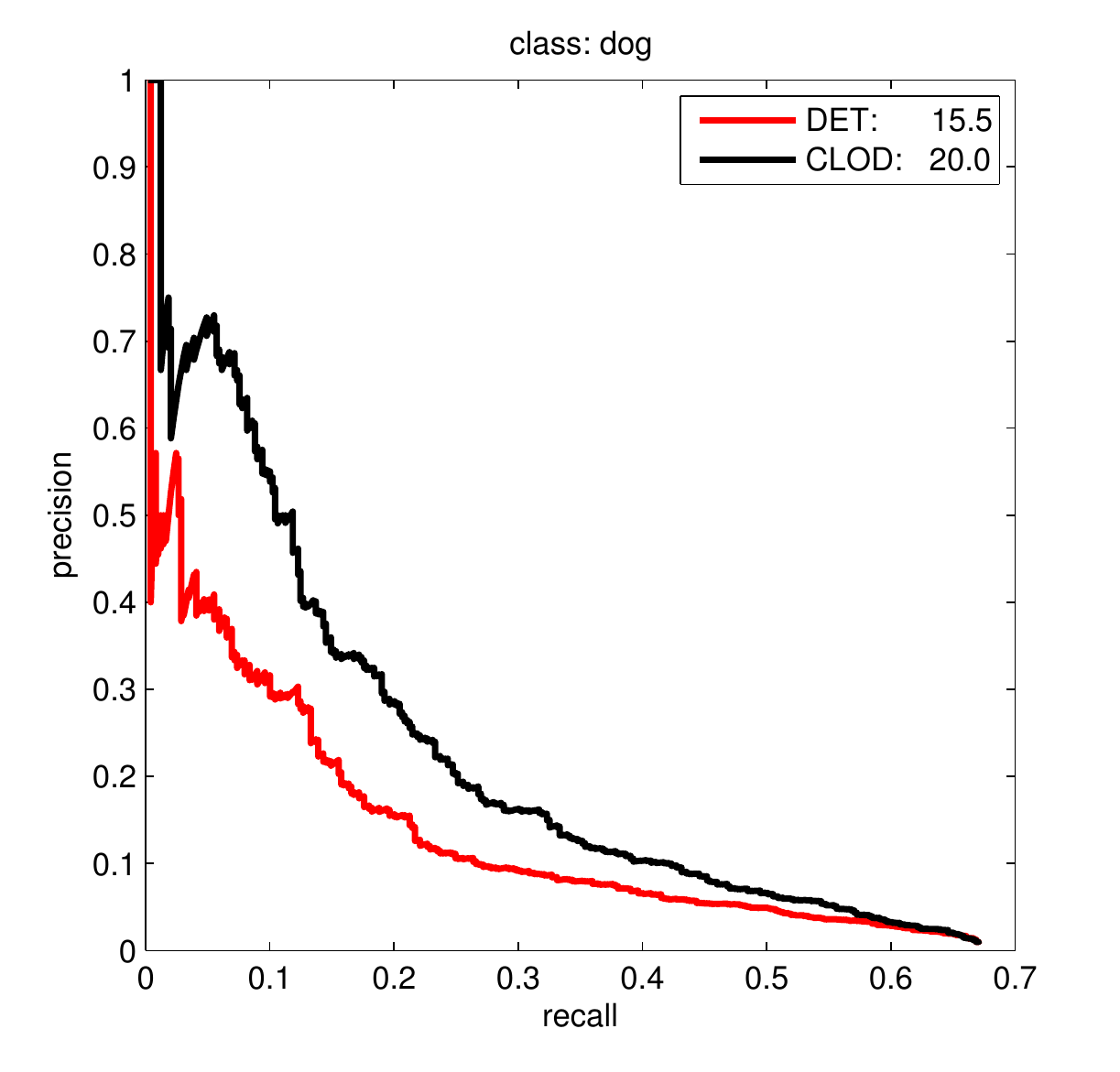}
  \includegraphics[width=35mm]{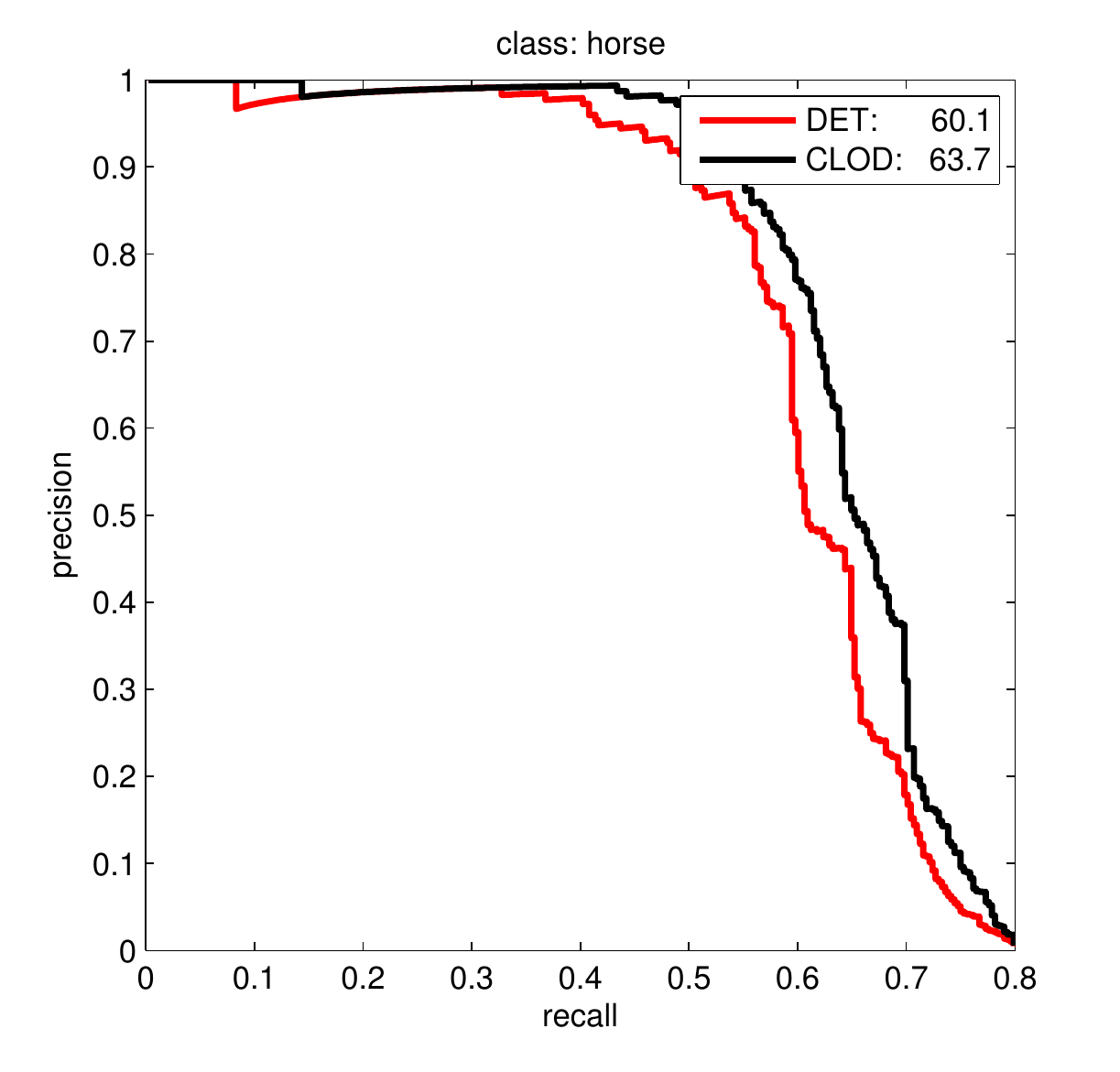}  
  \includegraphics[width=35mm]{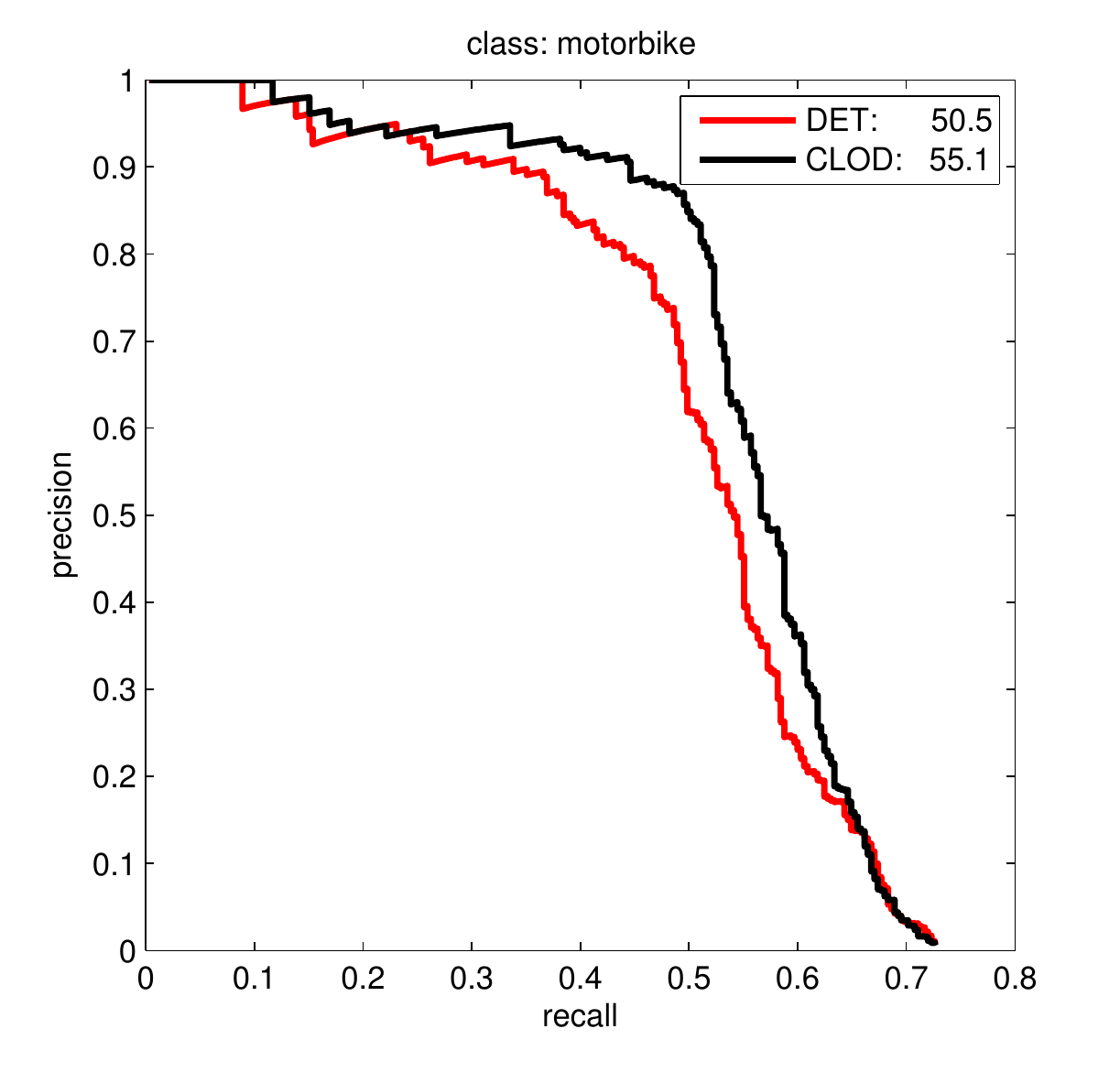}  
  \includegraphics[width=35mm]{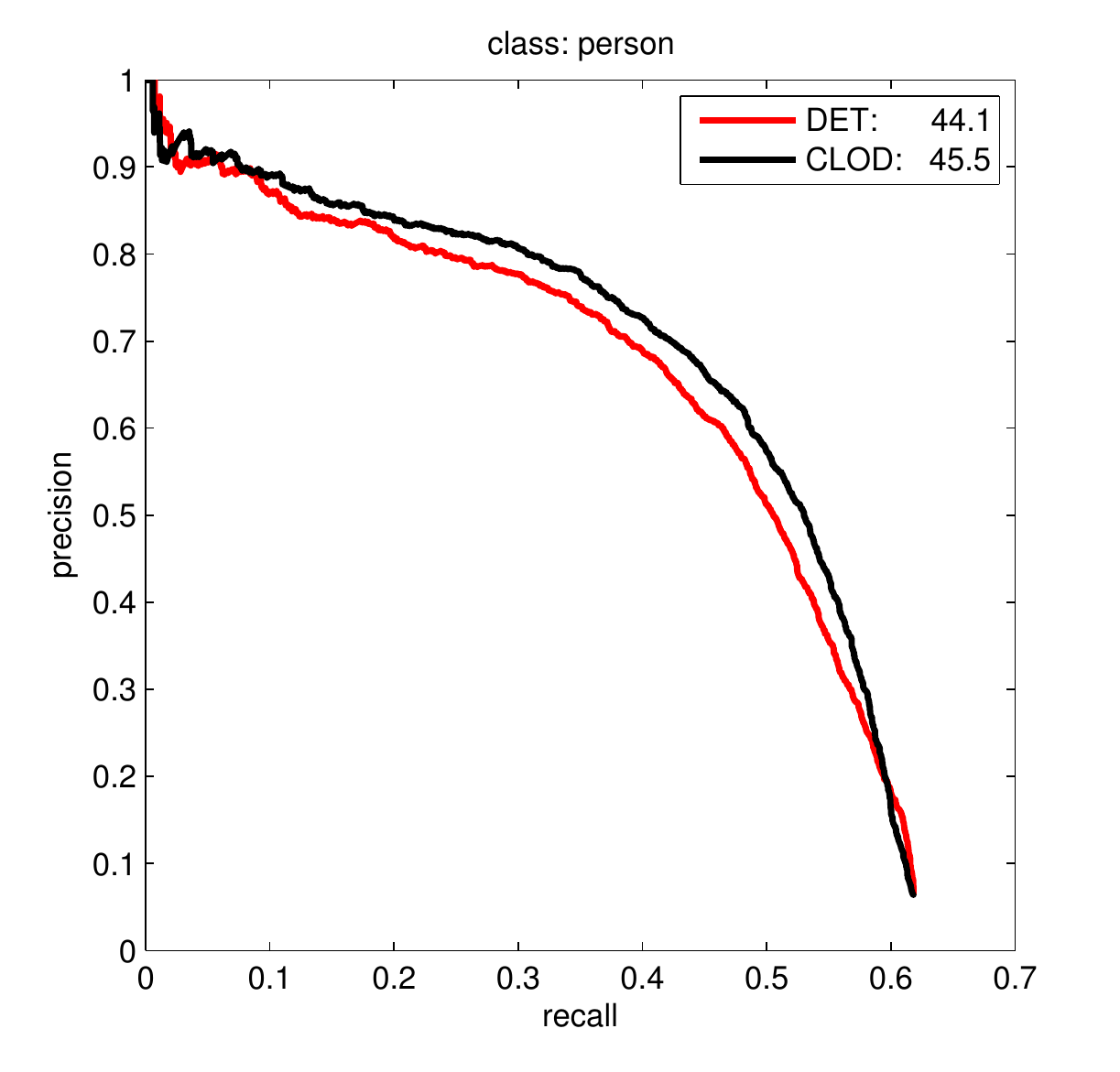}  \\  
  \includegraphics[width=35mm]{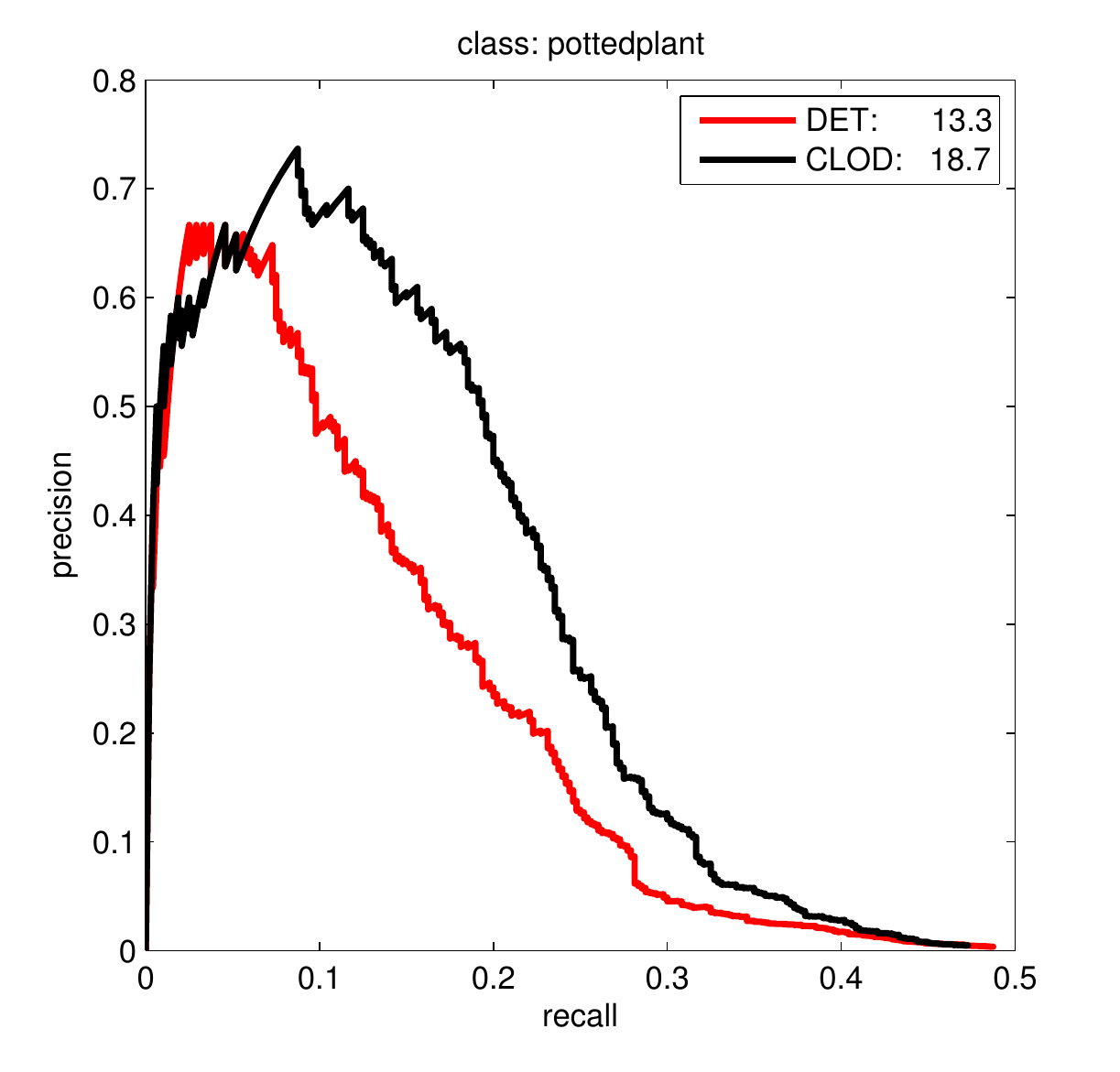} 
  \includegraphics[width=35mm]{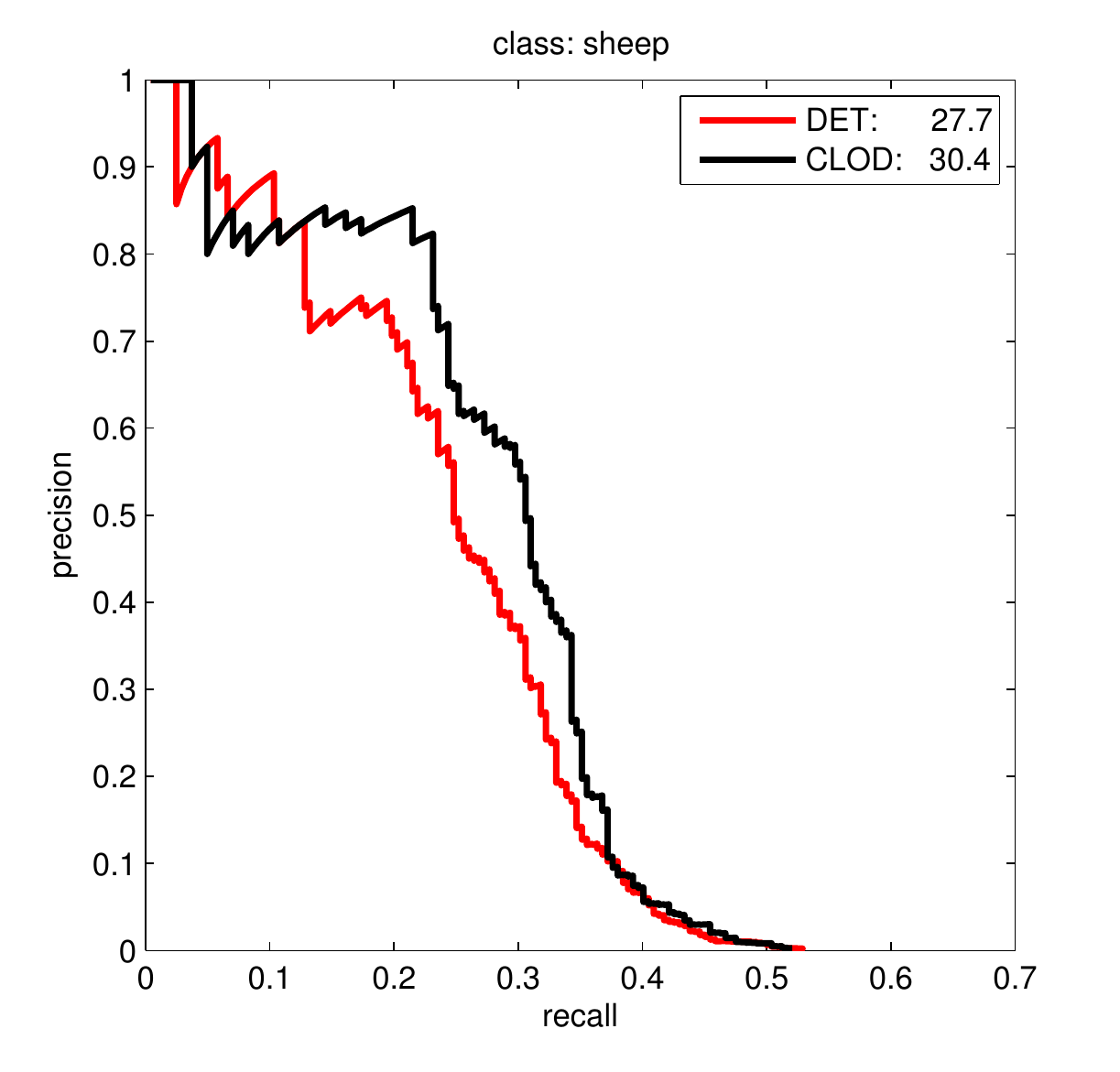}
  \includegraphics[width=35mm]{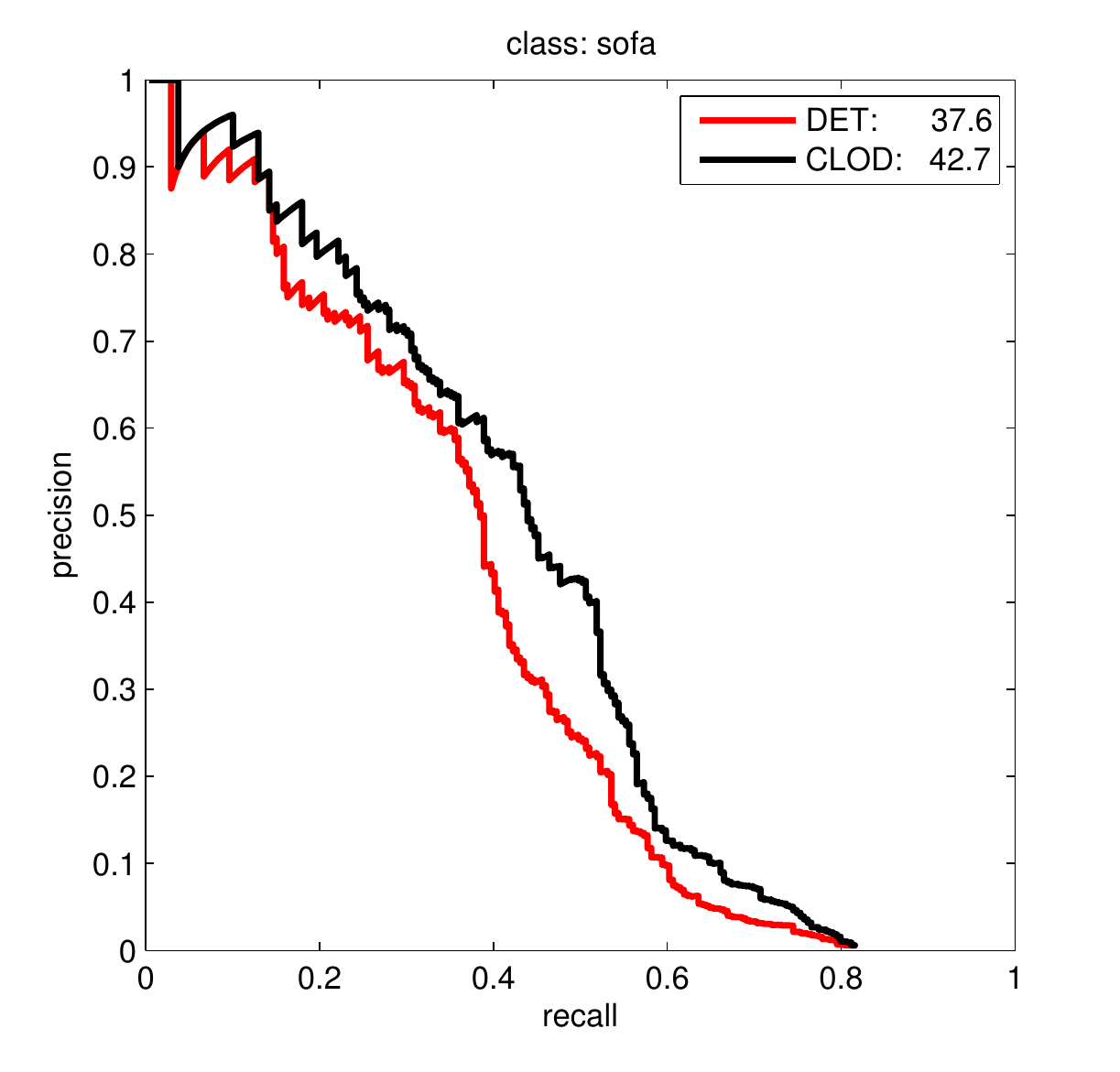}
  \includegraphics[width=35mm]{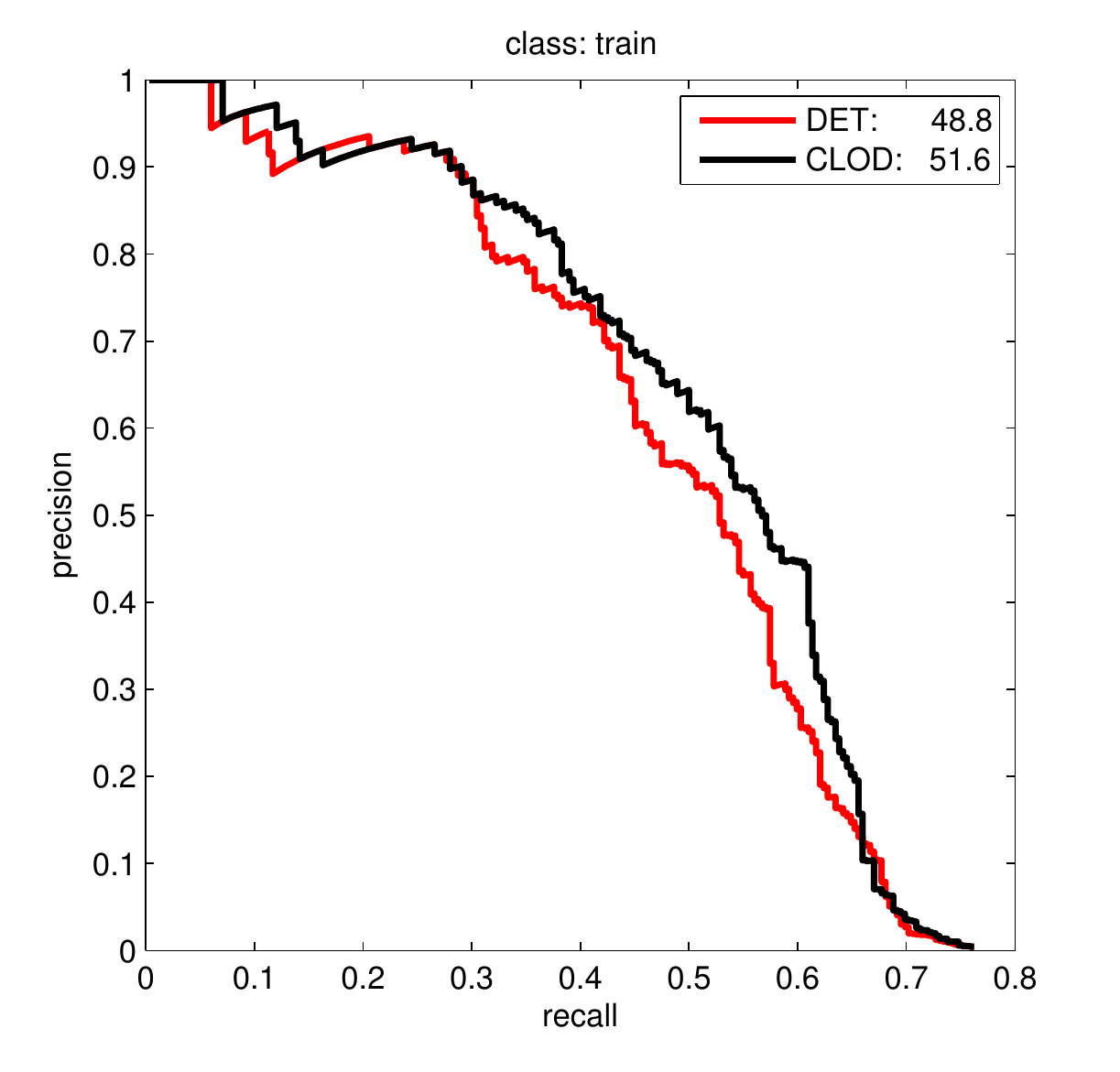}
  \includegraphics[width=35mm]{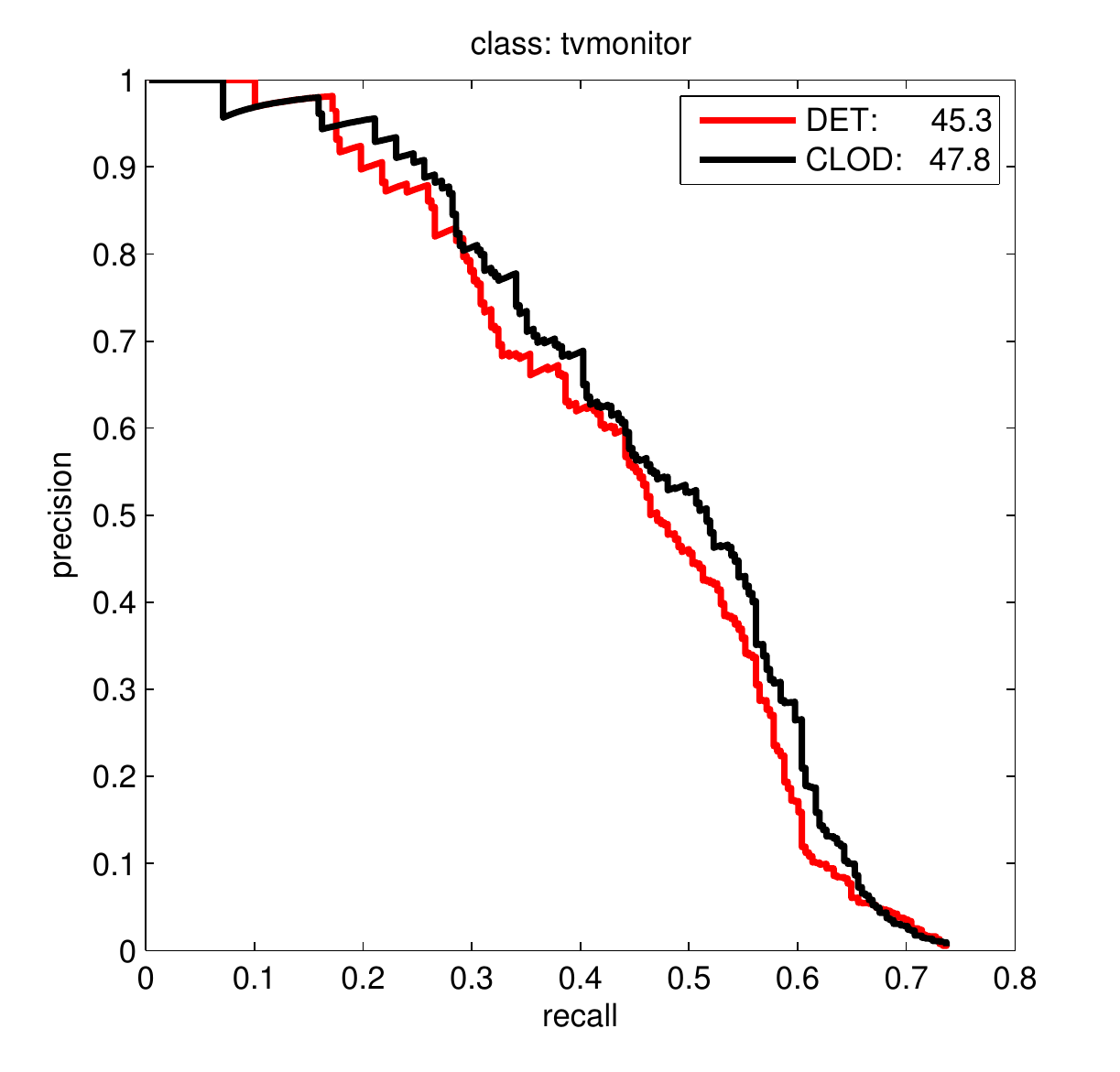} \\
  \end{tabular}
  \caption{Comparison of CLOD and original detection performance on PASCAL VOC 2007 detection dataset }\label{fig:ap }
\end{figure*}

Table~\ref{table:state_of_the_art} compares the CLOD with other state-of-the-art performance on PASCAL VOC 2007 detection dataset. The bold fond represents the first rank in related categories. Our methods achieved first place in 9 out of 20 categories , and rank first in mean average precision. The Figure~\ref{fig:ap } shows detailed precision-recall curve of CLOD and original detection algorithm over 20 categories in PASCAL VOC2007 detection dataset. The CLOD significantly improve the preliminary results for all categories.

\section{Conclusion}

In this paper, we have proposed a simple but powerful object detector called Classification Leveraged Object Detector. This detector needs a detection model and a classification model for each class. Extensive experiments on PASCAL2007 has shown the advantage of our approach. we achieved rank 1st for 9 categories and the mean AP is 39.5\%, which has achieved the state-of-the-art performance. 

\IEEEtriggeratref{25}

\bibliography{clod}
\bibliographystyle{IEEEtran}

\end{document}